\documentclass[sigconf]{acmart}
\usepackage{booktabs}

\usepackage{amssymb}
\usepackage{makecell}
\usepackage{float}
\usepackage{subcaption}
\usepackage[inkscapelatex=false]{svg}
\usepackage{pifont}
\usepackage{enumerate}
\usepackage{enumitem}
\usepackage{url}
\usepackage{newfloat}
\usepackage{listings}
\usepackage{multirow}
\usepackage{booktabs}
\usepackage{subcaption}
\usepackage[capitalize,noabbrev]{cleveref}
\usepackage{amsmath}
\usepackage{amssymb}
\usepackage{amsthm}
\usepackage{algorithm}
\usepackage{algpseudocode}
\usepackage{bm}
\usepackage{titlesec}

\newtheorem{remark}{Remark}
\AtBeginDocument{%
  }

\setcopyright{acmlicensed}
\copyrightyear{2025}
\acmYear{2025}
\setcopyright{acmlicensed}
\acmConference[KDD '25]{Proceedings of the 31st ACM SIGKDD Conference on Knowledge Discovery and Data Mining V.2}{August 3--7, 2025}{Toronto, ON, Canada}
\acmBooktitle{Proceedings of the 31st ACM SIGKDD Conference on Knowledge Discovery and Data Mining V.2 (KDD '25), August 3--7, 2025, Toronto, ON, Canada}
\acmDOI{10.1145/3711896.3737105}
\acmISBN{979-8-4007-1454-2/2025/08}

\usepackage{amsthm}

\begin{document}
\title{Revisiting Clustering of Neural Bandits:\\ Selective Reinitialization for Mitigating Loss of Plasticity}
\renewcommand{\shorttitle}{Revisiting Clustering of Neural Bandits:  Selective Reinitialization for Mitigating Loss of Plasticity}

\author{Zhiyuan Su}
\affiliation{%
  \institution{Gaoling School of Artificial Intelligence\\ Renmin University of China}
  \city{Beijing}
  \country{China}}
\email{ayinor0713@ruc.edu.cn}

\author{Sunhao Dai}

\affiliation{%
  \institution{Gaoling School of Artificial Intelligence\\ Renmin University of China}
  \city{Beijing}
  \country{China}}
\email{sunhaodai@ruc.edu.cn}

\author{Xiao Zhang}
\authornote{Xiao Zhang is the corresponding author.}
\affiliation{%
  \institution{Gaoling School of Artificial Intelligence\\ Renmin University of China}
  \city{Beijing}
  \country{China}}
\email{zhangx89@ruc.edu.cn}

\renewcommand{\shortauthors}{Zhiyuan Su, Sunhao Dai, and Xiao Zhang}

\begin{abstract}
Clustering of Bandits (CB) methods enhance sequential decision-making by grouping bandits into clusters based on similarity and incorporating cluster-level contextual information, demonstrating effectiveness and adaptability in applications like personalized streaming recommendations. However, when extending CB algorithms to their neural version (commonly referred to as Clustering of Neural Bandits, or CNB), they suffer from loss of plasticity, where neural network parameters become rigid and less adaptable over time, limiting their ability to adapt to non-stationary environments (e.g., dynamic user preferences in recommendation). To address this challenge, we propose Selective Reinitialization (SeRe), a novel bandit learning framework that dynamically preserves the adaptability of CNB algorithms in evolving environments. SeRe leverages a contribution utility metric to identify and selectively reset underutilized units, mitigating loss of plasticity while maintaining stable knowledge retention. Furthermore, when combining SeRe with CNB algorithms, the adaptive change detection mechanism adjusts the reinitialization frequency according to the degree of non-stationarity, ensuring effective adaptation without unnecessary resets. Theoretically, we prove that SeRe enables sublinear cumulative regret in piecewise-stationary environments, outperforming traditional CNB approaches in long-term performances. Extensive experiments on six real-world recommendation datasets demonstrate that SeRe-enhanced CNB algorithms can effectively mitigate the loss of plasticity with lower regrets, improving adaptability and robustness in dynamic settings.  
\end{abstract}



\maketitle

\section{Introduction}
\label{sec:intro}
Bandit methods have been widely used in streaming applications, such as streaming recommendation, but they typically assume that all decision recipients (e.g., users) share a single reward function, overlooking the potential similarities in their feedback mechanisms~\cite{li2010contextual,mehrotra2020bandit,Zhang2022Counteracting}. 
In many real-world applications, such as personalized recommendation and online advertising, user behaviors exhibit strong correlations, making a single reward function inefficient in balancing the exploration-exploitation trade-off~\cite{wang2023adcb, li2018online, nguyen2014dynamic, zhang2024reinforcing, zhang2023reward}.
To address this limitation, \textbf{C}lustering of \textbf{B}andits~(\textbf{CB}) methods have emerged as a powerful framework for enhancing sequential decision-making capabilities in streaming applications~\cite{gentile2014online,li2019improved, ban2021local, ban2024meta}. 
Taking recommendation as an example, CB algorithms treat each user as an individual bandit, dynamically grouping similar bandits into different user clusters based on their preferences. Within each user cluster, a distinct reward function is employed, enabling more effective collaborative information sharing and improving the modeling of user behavior heterogeneity. 
Considering the evolving nature of streaming applications, the goal of bandit methods is to minimize regret elegantly defined in various ways~\cite{yang2016tracking,zhang2018dynamic,fei2020dynamic,zhao2020simple,zhou2020nonstationary,ban2024meta,Zhang2022Counteracting}. 
The above advantages make CB particularly effective in large-scale systems with heterogeneous users, such as personalized recommendation~\cite{gentile2014online,ban2024meta}, dynamic pricing~\cite{miao2022context,wang2025dynamic}, and online advertising~\cite{ban2021local,gentile2017context}. 

However, most existing CB algorithms are \textbf{C}lustering of \textbf{L}inear \textbf{B}andits (\textbf{CLB}) methods, which assume a linear reward function~\cite{gentile2014online, li2019improved, ban2021local, gentile2017context, li2016collaborative, korda2016distributed, wang2024online, liu2022federated, zhang2025adao2b, zhang2021counterfactual}.  While many CLB algorithms offer strong theoretical guarantees and computational efficiency, their reliance on linearity limits their expressiveness in capturing complex user preferences. 
Some efforts have been made to provide neural extensions of traditional linear contextual bandit approaches~\cite{zhou2020neural,xu2020neural}, aiming to enhancing the representation capabilities of existing bandit methods, which has the potential to extend CLB into neural versions. 
A more recent approach, known as Meta Clustering of Neural Bandits~\cite{ban2024meta}, introduces a meta neural network to perform clustering, along with additional neural networks to estimate nonlinear rewards in the CB problem. 
However, as CB problems require dynamic clustering and the simultaneous online optimization of multiple neural networks to estimate distinct rewards in a bandit feedback setting, extending CLB to \textbf{C}lustering of \textbf{N}eural \textbf{B}andits (\textbf{CNB}) faces significant challenges, particularly regarding the convergence of the networks. 
Specifically, CNB algorithms encounter a critical issue known as ``loss of plasticity'', 
where neural network parameters struggle to train online on newly arrived data over time for model improvement (see Section~\ref{sec:loss_of_plasticity}), hindering their ability to adapt to users' evolving interests. This leads to increased cumulative regret of bandit policies and degraded performance in non-stationary environments.

For highly dynamic environments in bandit settings (e.g., streaming recommendation), simply retraining the entire neural network whenever user preferences shift is computationally infeasible~\cite{zhao2020simple, cheung2019learning}. A more efficient solution is to design a self-adaptive mechanism that selectively resets parts of the network to maintain its adaptability and plasticity.  
Intuitively, by periodically refreshing specific units instead of resetting the entire model, we can discard useless information while retaining useful parts of previously learned knowledge, thereby mitigating loss of plasticity in CNB. Next, we outline the key challenges of this idea and present our contributions in terms of approach, theoretical analysis, and evaluations. 

\textbf{Key Challenges.} To mitigate loss of plasticity in CNB, we face three key challenges.
(1) CNB algorithms rely on neural representations to incrementally estimate rewards for clustered bandits, but sustained training can cause unit inactivity and reduced adaptability. A principled approach is needed to identify and manage low-utility units without disrupting useful knowledge during bandit learning. Instead of random resets, an effective mechanism should quantify each unit's contribution on-the-fly and selectively reinitialize those with minimal impact.
(2) User preference shifts occur at varying rates across different scenarios, making it impractical to use a fixed reinitialization schedule~\cite{page1954continuous, hinkley1971inference}. An effective solution must dynamically adjust reinitialization frequency based on the degree of non-stationarity detected in the environment, preventing unnecessary resets while ensuring adaptation when significant shifts occur. 
(3) While reinitialization enhances adaptability, frequent resets can introduce high-variance predictions of bandits, harming cumulative regret in bandit optimization. Therefore, a robust mechanism should carefully balance stability and plasticity, ensuring that the model retains long-term knowledge while remaining flexible to evolving user behaviors. 

\textbf{Our Approach.} 
To address these challenges, we propose \textbf{Se}lective \textbf{Re}initialization (SeRe), a novel framework to restore neural network plasticity by intelligently refreshing only low-utility components, and design a change detection mechanism to embed the SeRe module into CNB algorithms.
Specifically, in each round, after carrying out the core CNB operations—such as receiving the target user, observing candidate arms, clustering users, updating neural networks, predicting rewards with confidence terms, and selecting the best arm—the algorithm plays the chosen arm and records the actual reward. It then employs a change detection mechanism to compare the observed and predicted rewards, adjusting a replacement rate parameter to gauge how significant any environmental shift is. Finally, the algorithm applies SeRe across all updated neural networks: each hidden layer updates the contribution of its units based on their activity and influence, and those identified as low in utility are selectively refreshed by reinitializing their weights.

\textbf{Theoretical Analysis.}  
We establish a theoretical foundation for SeRe by analyzing its regret in non-stationary environments. By selectively reinitializing low-utility units, SeRe prevents network stagnation while maintaining stable knowledge retention. We demonstrate that, under a piecewise-stationary assumption, SeRe enables CNB algorithms to achieve a \( \widetilde{\mathcal{O}}(\sqrt{T S}) \) regret bound, where \( S \) is the number of environment shifts. This ensures that SeRe-enhanced CNB algorithms remain adaptable while maintaining exploration-exploitation efficiency, even in highly dynamic settings.

\textbf{Evaluations.}
We validate SeRe through extensive experiments on six real-world recommendation datasets, integrating it into four state-of-the-art (SOTA) CNB algorithms (some of which are neural versions of CLB algorithms). Results show that SeRe reduces the cumulative dynamic regret by up to 12.82\% over 10,000 rounds while increasing the runtime by only a few milliseconds per round. Sensitivity analyses on key hyperparameters confirm SeRe’s robustness within a suitable parameter range, and plasticity analysis proves that our method can indeed solve the loss of plasticity of CNB. These findings demonstrate SeRe’s effectiveness in mitigating loss of plasticity for non-stationary environments, making it a practical and scalable solution for real-world streaming applications.

\section{Related Work}
\label{sec:related work}

\hspace*{1.1em}\textbf{Neural Bandits.} 
Traditional bandit methods assume linear reward functions~\cite{filippi2010parametric, soare2014best}, while neural bandits leverage deep networks for complex, non-linear reward structures. 
\citet{zhou2020neural} introduced a neural bandit framework based on Upper Confidence Bound (UCB) with theoretical guarantees, while \citet{zhang2020neural} used Thompson Sampling. \citet{xu2020neural} combined deep representation learning with UCB-based shallow exploration. Further improvements include perturbation-based exploration~\cite{jia2022learning}, active learning~\cite{ban2022improved, ban2024neural}, meta-learning~\cite{qi2024meta}, dual-network exploration~\cite{ban2021ee}, and federated approach~\cite{dai2022federated}. At the application level, many studies have focused on recommender systems employing neural bandits for adaptive preference tracking~\cite{santana2020contextual, shen2023hyperbandit, Zhang2022Counteracting}. Additionally, several works have explored variants and sub-problems of neural bandits~\cite{kassraie2022neural, hong2020latent, maillard2014latent}. However, these works often use only a single bandit, limiting the scalability and adaptability.

\textbf{Clustering of Bandits.}  
Clustering techniques improve bandit efficiency by grouping similar users. CLUB~\cite{gentile2014online} pioneered similarity-based clustering, extended by SCLUB~\cite{li2019improved} with dynamic user merging and splitting. LOCB~\cite{ban2021local} introduced local clustering for overlapping user groups, while other works explored feature-based clustering~\cite{gentile2017context, li2016collaborative}, distributed settings~\cite{korda2016distributed}, online clustering~\cite{wang2024online}, and federated learning~\cite{liu2022federated}. However, most methods assume linear reward functions, limiting adaptability to complex user preferences. M-CNB~\cite{ban2024meta} incorporated neural networks and meta-learning to overcome this, yet it does not explicitly address loss of plasticity, making it less effective in non-stationary environments.

\textbf{Continual Learning.}  
Continual learning primarily focuses on two problems. 
The first is catastrophic forgetting, which has been addressed by methods such as pseudo-rehearsal~\cite{robins1995catastrophic}, EWC~\cite{kirkpatrick2017overcoming}, GEM~\cite{lopez2017gradient}, Deep Generative Replay~\cite{shin2017continual}, and geometric approaches~\cite{chaudhry2018riemannian}. 
The second is adaptation to non-stationary environments, for which approaches including weight adjustments~\cite{adel2019continual}, Bayesian methods~\cite{kurle2019continual}, variational inference~\cite{rudner2022continual}, meta-learning~\cite{wang2022meta}, incremental learning~\cite{liu2020incremental}, and natural gradient-based strategies~\cite{tseran2018natural} have been proposed.
Moreover, the fact that model plasticity degrades over time~\cite{ellis2000age,van2022three, sokar2023dormant, abbas2023loss} prompts solutions such as Regenerative Regularization~\cite{kumar2023maintaining} and Continual Backpropagation~\cite{dohare2023maintaining, dohare2024loss}.

\section{Problem Formulation and Analysis} 
This section defines the CNB problem in recommendation scenarios, and analyzes the loss of plasticity in CNB.

\subsection{Clustering of Neural Bandits (CNB) Problem}
\label{sec:problem}

First, we introduce the Clustering of Neural Bandits  problem, which is characterized by incorporating the correlations between bandits with different reward functions into the decision-making process of arm selection at each round~\cite{ban2024meta}. To provide a more concrete illustration of the CNB problem, we frame it within the context of bandit-based recommender systems.

\textbf{Bandit-based Recommendation.} We consider a personalized recommender system involving \( n \) bandits (i.e., \( n \) users), represented by the user set \( \mathcal{N} = \{1, 2, \dots, n\} \), which interacts with a platform over multiple rounds. At the \( t \)-th round, a user \( u_t \in \mathcal{N} \) logs into the platform, and the platform then recalls a set of \( K \) candidate arms (i.e., \( K \) candidate items, e.g., products in e-commerce) from the arm pool $\mathcal{A}$, denoted by \( \mathcal{A}_t = \{\bm{a}_t^1, \bm{a}_t^2, \dots, \bm{a}_t^K\} \). Each arm \( \bm{a}_t^i \in \mathbb{R}^d \) (where \( i \in [K] := \{1, 2, \dots, K\} \)) is represented by a \( d \)-dimensional feature vector, capturing current user context and item-specific characteristics~\cite{li2010contextual}. The platform further selects an arm \( \bm{a}_t^{I_t} \in \mathcal{A}_t  \) to recommend to user \( u_t \), and the user \( u_t \) responds to the recommended item \( \bm{a}_t^{I_t}\) by providing feedback in the form of a reward \( r^{I_t}_t | u_t \) (corresponding to user behaviors such as clicks and conversions), where \( I_t \in [K] \) denotes the index of the arm selected by the bandit policy at round \( t \).
For all $i \in [K]$, the reward \( r^i_t | u_t \) reflects the user's preference for the recommended arm \(\bm{a}_t^{i}\). 
Specifically, the reward generated by user \( u_t \) for arm \( \bm{a}_t^i \) is modeled as:
\begin{equation}
r_t^i | u_t = g_{u_t,t}(\bm{a}_t^i) + \xi_t^i,
\label{eq:reward_function}
\end{equation}
where \( g_{u_t, t} \) is the \emph{true reward} 
of user \( u_t \) at round \( t \), which maps arm \( \bm{a}_t^i \) to a reward, and \( \xi_t^i \) is a noise term with zero mean, i.e., \( \mathbb{E}[\xi_t^i] = 0 \). Additionally, as in existing works~\cite{ban2021local, gentile2014online}, we assume that the reward \( r_t^i \in [0, 1] \). Note that the platform never knows the true \( g \), but instead uses neural networks to learn and represent the reward mapping of the users.

The goal of the platform is to minimize the \emph{cumulative dynamic regret} over \(T\) rounds, defined as:
\begin{equation}
\label{eq:regret}
\mathbf{R}_T = \sum_{t=1}^T \big[g_{u_t,t}(\bm{a}_t^*) - g_{u_t,t}(\bm{a}^{I_t}_t) \mid u_t, \mathcal{A}_t  \big],
\end{equation}
where \(\bm{a}_t^* \in \mathcal{A}_t \) represents the arm that maximizes \(g_{u_t,t}(\bm{a})\) at round \(t\), i.e., \(
\bm{a}_t^* = \arg\max_{\bm{a} \in \mathcal{A}_t } g_{u_t,t}(\bm{a})\).

\textbf{Clustering.} 
Another important module in the CNB problem is clustering. Users (each user corresponds to one bandit) are often grouped into different clusters, where users within each cluster have similar preferences or exhibit similar behaviors. Therefore, the key point of clustering is to put users with sufficiently similar preferences into the same cluster, and users with insufficiently similar preferences into different clusters. We define clustering as:
\begin{definition}[$(\epsilon_1, \epsilon_2)$-User Cluster]
\label{def:cluster}
At round $t$, given a metric $\mathcal{M}$ that measures the user preferences, for an arm $\bm{a}^i_t \in \mathcal{A}_t $, an $(\epsilon_1, \epsilon_2)$-user  cluster $\mathcal{C}(\bm{a}^i_t) \subseteq \mathcal{N}$ with respect to $\bm{a}^i_t$ satisfies:
\begin{enumerate}
    \item $\forall u, u' \in \mathcal{C}(\bm{a}^i_t)$, there exists a constant $\epsilon_1>0$, such that $\|\mathcal{M}(u|\bm{a}^i_t)-\mathcal{M}(u'|\bm{a}^i_t)\| \leq \epsilon_1$.
    \item $\nexists \mathcal{C}' \subseteq \mathcal{N}$, s.t. $\mathcal{C}'$ satisfies (1) and $\mathcal{C}(\bm{a}^i_t) \subset \mathcal{C}'$.
    \item Given two different clusters $\mathcal{C}(\bm{a}^i_t)$ and $\mathcal{C}'(\bm{a}^i_t)$, there exists a constant $\epsilon_2 > 0$, such that $\forall u \in \mathcal{C}(\bm{a}^i_t), u' \in \mathcal{C}'(\bm{a}^i_t), \|\mathcal{M}(u|\bm{a}^i_t)-\mathcal{M}(u'|\bm{a}^i_t)\| \geq \epsilon_2$.
\end{enumerate}
\end{definition}

\begin{figure}[t]
    \centering
    \includegraphics[width=1\linewidth]{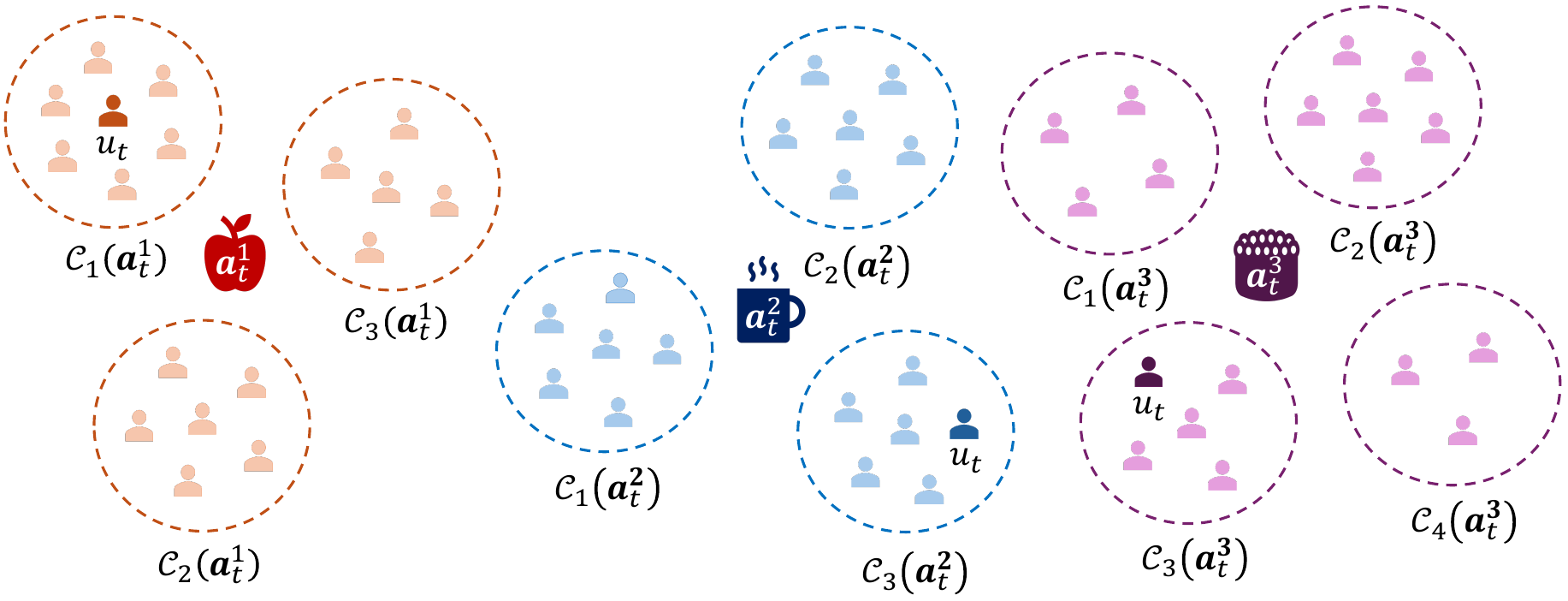}
    \caption{Schematic illustration of clustering. For each item, users are grouped into clusters based on similarity in preferences or behaviors. The figure reflects how clusters adapt to specific items, illustrating item-varying user pieceation.}
    \label{fig:clustering}
\end{figure}

The above definition shows that the clusters we consider vary from item to item, which is quite reasonable since users with similar preferences for one product may exhibit different preferences for another. Given item $\bm{a}^i_t$, the $\mathcal{M}$ differences of users within a cluster are limited, while those between clusters are obvious~\cite{ban2021local,gentile2014online,gentile2017context,li2016collaborative,li2019improved}. Let $q^i_t$ be the number of clusters given arm $\bm{a}^i_t$, where \(q^i_t \ll n\).
In existing studies, there are different ways to characterize $\mathcal{M}$.
In earlier studies exploring CLB algorithms~\cite{ban2021local,gentile2014online,li2019improved}, \(\mathcal{M}\) is often assumed to be independent of items (i.e., for any two items \(\bm{a}, \bm{a}' \in \mathcal{A}_t\), \(\mathcal{M}(u|\bm{a})=\mathcal{M}(u|\bm{a}')\)) and is represented by the user's linear bandit parameter. The clustering process is based on the differences (usually measured by the \(\ell_2\)-norm) between these parameters, under the assumption that users with similar bandit parameters should belong to the same cluster. Moreover, since \(\mathcal{M}\) is assumed to be invariant across items in these studies, $q^{1}_t=q^{2}_t=\dots=q^{K}_t$ and let them be equal to $q_t$. 
For any \(\bm{a}, \bm{a}' \in \mathcal{A}_t\), $\mathcal{C}_j(\bm{a})$ is exactly the same as $\mathcal{C}_j(\bm{a}')$ for all \(j \in \{1,2,\dots,q_t\}\).
However, in the CNB algorithms we focus on, \(\mathcal{M}\) is defined based on the expected reward for each item. In the M-CNB algorithm~\cite{ban2024meta}, for example, the threshold \(\epsilon_1\) is strictly set to zero to ensure that users within the same cluster have identical reward expectations for a given item, while \(\epsilon_2\) is a positive constant. Consequently, for a specific item \(\bm{a}^i_t\), the user set \(\mathcal{N}\) is divided into \(q^i_t\) clusters, \(\mathcal{C}_1(\bm{a}^i_t), \dots, \mathcal{C}_{q^i_t}(\bm{a}^i_t)\). The detailed schematic illustration of clustering is shown in Figure \ref{fig:clustering}.

\textbf{More Notations.} 
To facilitate modeling and computation, all existing CNB frameworks employ UCB-based arm selection policy. At each round \(t\) for user \(u_t\), the algorithm considers a set of candidate arms \(\{\bm{a}_t^1, \bm{a}_t^2, \dots, \bm{a}_t^K\}\) and computes, for each arm, an upper confidence value
$
U^i_t = \hat{r}^i_t + \text{Confidence Term}
$, 
where \(\hat{r}^i_t\) is the predicted reward for arm \(\bm{a}_t^i\). \(\hat{r}^i_t\) is estimated only by the user learner, while the confidence term is generated by a general learner leveraging both user-level information (obtained from a user network) and cluster-level information (derived from a network representing the corresponding user cluster) to compute. In each round, the learners are continuously updated.

\subsection{Loss of Plasticity in CNB}
\label{sec:loss_of_plasticity}
The CNB framework shows significant potential in recommender systems by balancing exploration and exploitation through neural networks. However, existing CNB algorithms face a key challenge, \emph{loss of plasticity}, limiting adaptability in dynamic environments~\cite{dohare2024loss}. This section analyzes the motivation and feasibility of extending CLB to CNB, and then discusses the loss of plasticity issue.

\textbf{Necessity and Feasibility of Neural Extension.} 
Most existing CB algorithms are CLB algorithms (e.g., CLUB~\cite{gentile2014online}, SCLUB~\cite{li2019improved}, LOCB~\cite{ban2021local}). However, real-world recommender systems often exhibit complex, non-linear user preferences~\cite{ban2024meta, zhou2020neural, valko2013finite}, making the linear assumption too restrictive. To overcome this, CNB algorithms replace linear mappings with deep neural networks that have universal approximation capabilities~\cite{cybenko1989approximation}. Direct adaptation of CLB frameworks to neural architectures is challenging because many clustering mechanisms depend on explicit parameter similarity metrics that do not extend naturally to high-dimensional spaces. A practical solution is the \textit{deep representation, shallow exploration} paradigm~\cite{xu2020neural}, in which deep networks extract latent features while exploration policies such as UCB or Thompson Sampling are applied on the final layer. In this work, we extend CLUB, SCLUB, and LOCB into their neural counterparts—denoted as CLUB-N, SCLUB-N, and LOCB-N, respectively.

\textbf{Loss of Plasticity in CNB.} 
Loss of plasticity in CNB algorithms means that over time, neural networks tend to overfit to past data, causing weight updates to reinforce existing patterns while reducing flexibility to accommodate new changes. This results in an increase in inactive units (e.g., dead ReLU neurons), an expansion in weight magnitudes, and a decline in effective rank, ultimately degrading the model’s ability to learn evolving user preferences.

\begin{figure}[t]
    \centering
    \includegraphics[width=1.0\linewidth]{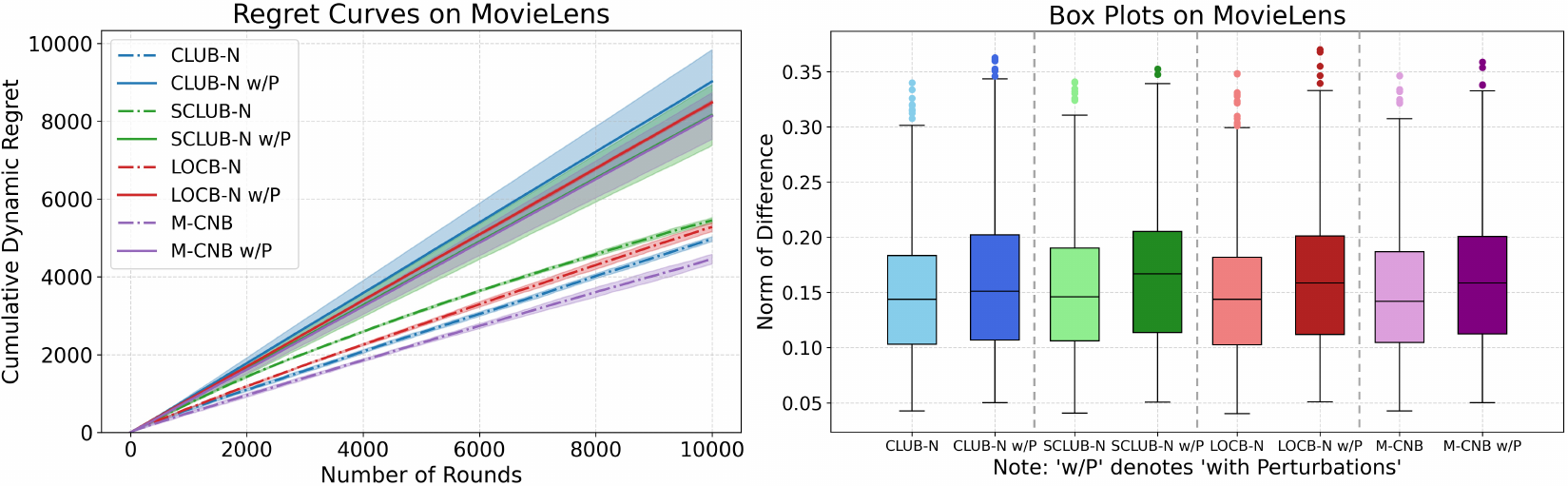}
    \caption{Loss of plasticity in existing CNB algorithms. (1)~The left panel: the ``-N'' suffix indicates the neural version of the method, and ``w/P'' (i.e. ``with Perturbations'') means that periodic perturbations are added to the user features. Five experiments were performed for each setting: the middle line represents the average curve and the shaded area represents the 95\% confidence interval. (2)~The right panel: this box plot illustrates the $\ell_2$-norm of the difference in the last layer's parameters, computed from samples taken every 25 rounds over 10,000 rounds on the MovieLens dataset.}
    \label{fig:lossofplasticity}
\end{figure}

To empirically validate this issue, we conducted experiments using the MovieLens dataset. We conducted experiments on the MovieLens dataset comparing CNB algorithms under relatively stable user features versus scenarios with small, periodic perturbations. As a recommendation dataset, MovieLens inherently exhibits some temporal variation. However, to more explicitly observe CNB's performance degradation in non-stationary environments, we introduced periodic perturbations to both user and item features.
Specifically, every 200 rounds we added Gaussian noise (\(\mathcal{N}(0, 0.1^2)\)) to user features to simulate gradual preference shifts.
As shown in the left panel of Figure~\ref{fig:lossofplasticity}, while CNB algorithms exhibit sublinear regret under stable conditions, the cumulative regret increases almost linearly with perturbations, indicating a severe loss of plasticity.

In the right panel of Figure~\ref{fig:lossofplasticity}, we present a box plot of the $\ell_2$-norm of the difference in the updated parameters of the neural network's last layer, computed every 25 rounds over 10,000 rounds of MovieLens dataset across eight experimental sets.
Since perturbations occur every 200 rounds, one expects abrupt changes at those multiples. However, in the four groups of experiments, the median of the norm is not much different in the absence of perturbations (left side of each group) and with perturbations (right side of each group), and the maximum of difference is only about 0.02; and if we compare the difference between the maximum values of the two figures in each group, it is only around 0.03. This indicates that the network parameters become relatively rigid and hinder adaptation in non-stationary scenarios. This finding emphasizes the necessity of a mechanism to maintain neural plasticity in CNB algorithms. 

\section{SeRe: The Proposed Approach}
\label{sec:SeRe}
In the previous section, we identified that CNB algorithms suffer from a loss of plasticity in non-stationary environments, as neural networks become overly adapted to historical interactions and thus struggle to learn new patterns. The challenge is to preserve model flexibility without discarding useful accumulated knowledge. Although reinitializing parameters is a natural idea, naïve approaches, such as reinitializing all or randomly selected units~\cite{zhao2020simple}, do not perform well in CNB due to the large number of networks and moderate environmental changes.

Motivated by recent studies that selectively reinitialize network parameters by retaining high-contribution units and replacing those with low contribution~\cite{dohare2023maintaining, dohare2024loss}, we propose \textbf{Se}lective \textbf{Re}initialization, named SeRe. SeRe, described in detail in Section \ref{sec:sere_details}, is a continual 
bandit learning approach designed to enhance CNB algorithms in non-stationary settings. It is built on two key mechanisms: a \emph{contribution utility update}, which quantifies each unit's importance based on its activation and outgoing weight magnitudes, and a \emph{parameter reinitialization} procedure that reinitializes underutilized units. In Section \ref{sec:sere_app}, to further integrate SeRe into CNB, we add an adaptive \emph{change detection} module that dynamically adjusts the reinitialization frequency based on shifts in user preferences.

In the remainder of this section, we detail the design of SeRe and discuss its application within CNB algorithms.

\subsection{Details of SeRe}
\label{sec:sere_details}
Next, we will first introduce the two key mechanisms of SeRe separately, and then present the complete SeRe algorithm. Before introducing the specific mechanism, we need to clarify some notations. For the neural networks in CNB algorithms, we do not consider the last layers. Let \(h_{l, i}\) denote the output of the \(i\)-th hidden unit in the \(l\)-th layer and \(w_{l, i, j}\) denote the weight connecting the \(i\)th unit in the \(l\)-th layer and the \(j\)-th unit in the \((l+1)\)-th layer. For the \(i\)-th unit in layer \(l\), we define the contribution utility as \(u_{l, i}\). In addition, the age of the unit \(\textit{age}_{l, i}\) represents the number of steps since the last reinitialization. We also define a counter \(c_l\) for each layer \(l\) as a control parameter for reinitialization. At the beginning, we initialize all \(u_{l, i}\), $\textit{age}_{l, i}$ , $c_l$ to zero.

There are three paragraphs in this section. ``Contribution utility update'' describes how to quantify and update the contribution of each hidden unit based on its activation and its outgoing weights, while ``Parameter Reinitialization'' details how to selectively refresh units with low utility and sufficient age. In ``Integration of Components'', these components are integrated to ensure that the network can continuously adapt to new data while retaining valuable learned information.

\textbf{Contribution Utility Update.}
Contribution utility update plays a crucial role in maintaining the adaptability of the network in SeRe. 
The goal is to quantify how much each hidden unit contributes to decision-making by considering both its activation and the strength of its outgoing weights. As the model learns from evolving data, units' relevance can change—units with weak activations or small outgoing weights contribute little and may hinder the network's adaption.  
To identify such units, we dynamically compute each unit's contribution utility as the sum of the utilities of all its outgoing connections, providing a comprehensive measure of its relevance to the network’s output. Specifically, the sum of the products of a unit’s activation and the absolute value of its outgoing weights reflects its impact on subsequent layers. If a unit's contribution is low relative to others, it is deemed low-utility~\cite{dohare2023maintaining, dohare2024loss}. 
To ensure the utility measure adapts to recent changes, we adopt a running average with a decay factor. This update mechanism assigns more weight to recent contributions while still considering past information, preventing outdated values from unduly influencing the reinitialization process.

The contribution utility is updated iteratively at each step as:
\begin{equation}
    u_{l, i} \leftarrow \eta \cdot u_{l, i} + (1 - \eta) \cdot |h_{l, i}| \cdot \sum_{j=1}^{n_{l+1}} \big|w_{l, i, j} \big|,
    \label{eq:contribution_utility_time_agnostic}
\end{equation}
where $\eta \in [0, 1]$ is the decay rate that controls the balance between historical and current contributions. By using this design, we effectively identify units that are no longer contributing meaningfully to the network and can prioritize them for reinitialization, ensuring the network's continued ability to adapt to new data and maintain flexibility in decision-making.

\textbf{Parameter Reinitialization.} 
Parameter reinitialization in SeRe is essential for maintaining network adaptability by replacing outdated or dormant units. As user preferences evolve, some units become less effective, hindering learning. To address this, SeRe identifies units for reinitialization based on their contribution utility and age. 
A unit is eligible if its age exceeds a predefined maturity threshold \(m\), and reinitialization is triggered when the counter \(c_l\) for layer \(l\) reaches or exceeds 1. Here, the replacement rate \(\rho\) controls how quickly \(c_l\) accumulates by scaling \(s_m\) (i.e., the number of units whose age is greater than \(m\)) thereby adjusting the reinitialization frequency. The unit with the lowest contribution utility among eligible candidates is selected. Its input weights are reinitialized from a predefined distribution \(\mathcal{D}\) to promote diversity and enable fresh representation learning, while its output weights are reset to zero to avoid inherited biases and abrupt interference. 
Here we empirically use Kaiming Uniform Initialization~\cite{he2015delving}, specifying the distribution \(\mathcal{D} = \mathcal{U}\left(-\sqrt{\frac{6}{n_{\text{in}}}}, \sqrt{\frac{6}{n_{\text{in}}}}\right)\), 
where \(n_{\text{in}}\) is the number of input connections to the unit, because our neural network primarily uses ReLU activation function, and Kaiming Uniform Initialization helps maintain stable signal variance through the layers. The unit's contribution utility and age are reset to zero, and \(c_l\) is decremented accordingly.
This mechanism prevents the accumulation of stale parameters, allowing the network to dynamically adjust to evolving data while maintaining stability and efficiency.

\begin{algorithm}[t]
\caption{SeRe ($f$)}
\label{alg:SeRe}
\begin{algorithmic}[1]
\State \textbf{Input:} 
\(\rho\) (replacement rate),
\(\eta\) (decay rate),
\(m\) (maturity threshold), 
\(\mathcal{D}\) (initial weights distribution)
\For{each layer \(l\) (except for the last layer)}
    \For{each unit \(i\) in layer \(l\)}
        \State \parbox[t]{\dimexpr\linewidth-\algorithmicindent}{Update \(u_{l, i}\) using Update Equation \eqref{eq:contribution_utility_time_agnostic}}
        \State Increment the age of the unit: \(\textit{age}_{l, i} \leftarrow \textit{age}_{l, i} + 1\)
    \EndFor
    \State \parbox[t]{\dimexpr\linewidth-\algorithmicindent}{Find \(s_{m} \leftarrow\) number of units with age greater than $m$}
    \State \parbox[t]{\dimexpr\linewidth-\algorithmicindent}{
    Update counter: \(c_l \leftarrow c_l+\rho \cdot s_{m}\)}
    \If{\(c_l \geq 1\)}
        \State \parbox[t]{\dimexpr\linewidth-\algorithmicindent}{Select the unit \(\tilde{i} = \arg\min_{i \in \{j | \textit{age}_{l, j} > m\}} u_{l, i}\)}
        \State \parbox[t]{\dimexpr\linewidth-\algorithmicindent}{Reinitialize input weights:\\ \( w_{l-1, k, \tilde{i}} \sim \mathcal{D}, \quad \forall k \in \{1, \dots, n_{l-1}\} \)}
        \State \parbox[t]{\dimexpr\linewidth-\algorithmicindent}{Reinitialize output weights to zero:\\ \( w_{l+1, \tilde{i}, j} \leftarrow 0, \quad \forall j \in \{1, \dots, n_{l+1}\} \)}
        \State \parbox[t]{\dimexpr\linewidth-\algorithmicindent}{Update utility, age, and counter:\\ \( u_{l, \tilde{i}} \leftarrow 0, \quad \textit{age}_{l, \tilde{i}} \leftarrow 0, \quad c_l \leftarrow c_l - 1 \)}
    \EndIf
\EndFor
\end{algorithmic}
\end{algorithm}

\begin{figure}[!t]
    \centering
    \includegraphics[width=1.0\linewidth]
    {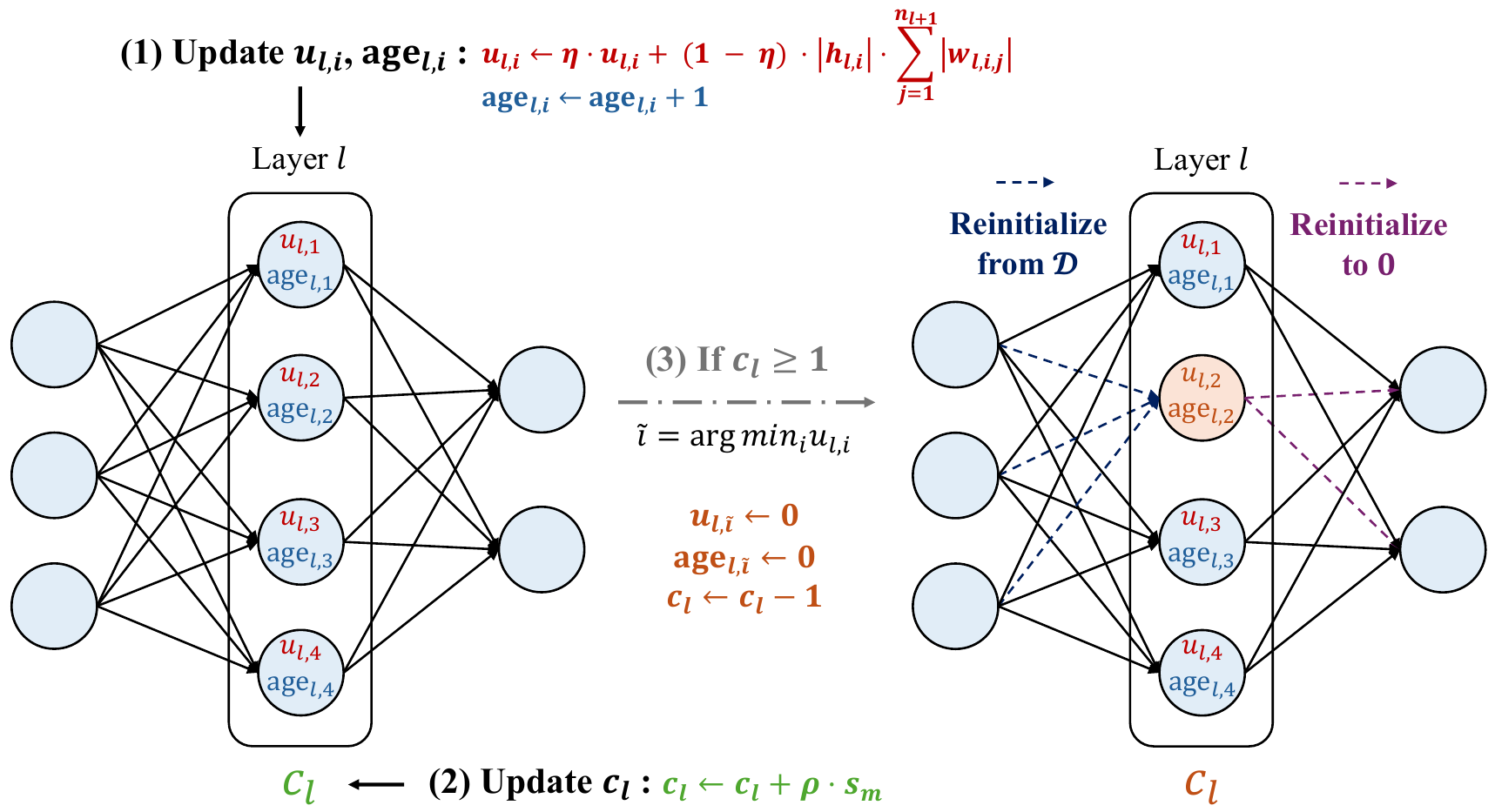}
    \caption{SeRe workflow at layer \( l \): (1) Update contribution utility \( u_{l,i} \) and age \( \textit{age}_{l,i} \). (2) Increment counter \( c_l \) based on matured units \( s_m \). (3) If \( c_l \geq 1 \), then the unit with the lowest utility and the weights associated with it are reinitialized and the metric is updated.}
    \label{fig:sere}
\end{figure}

\textbf{Integration of Components.}  
SeRe integrates the above two mechanisms to maintain the plasticity of the neural network, as shown in Algorithm \ref{alg:SeRe}. 
It first updates contribution utility for all units and increments their age, identifying low-utility units. Units exceeding the maturity threshold \(m\) contribute to updating the reinitialization counter \(c_l\). If \(c_l \geq 1\), the lowest-utility unit among those exceeding \(m\) is selected for reinitialization. Its input weights are resampled from \(\mathcal{D}\) for adaptability, while output weights are reset to zero to prevent residual influence. The unit’s utility and age are reset, and the reinitialization counter is updated. This targeted approach preserves network plasticity while ensuring stability and efficiency. 
For a more intuitive presentation, we show the SeRe workflow at each layer in Figure \ref{fig:sere}.

\subsection{Application of SeRe in CNB Algorithms} 
\label{sec:sere_app}
In SeRe, $\rho$ is a very important parameter that controls the speed at which $c_l$ is updated and thus the frequency of reinitialization.
To dynamically adjust the replacement rate \(\rho\) in response to environmental shifts, we maintain a change detection statistic using the Page-Hinkley (PH) method~\cite{page1954continuous, hinkley1971inference}, a variant of the Cumulative Sum (CUSUM) test. 
Unlike the standard PH test which tracks directional drift, we introduce an absolute-value variant, denoted as PHA, i.e., Page-Hinkley-Absolute. This design accumulates the absolute prediction error, ensuring that we detect both underestimation and overestimation of rewards, aligning better with our goal of monitoring model prediction performance regardless of direction.
At round \(t\), for user \(u_t\), the platform selects an arm in \(\mathcal{A}_t \). Let \(r_t\) be the actual reward at round ~\(t\), and \(\hat{r}_t\) be the model’s predicted reward for the arm. We define:
\begin{equation}
    \mathrm{PHA}_t \;=\; \mathrm{PHA}_{t-1} \;+\; \bigl[\,|r^{I_t}_t - \hat{r}^{I_t}_t|\;-\;\delta \bigr],
\end{equation}
where \(\delta > 0\) is a small offset controlling the sensitivity of detection. We also keep track of
\(\mathrm{PHA}_{\min} = \min_{k \le t}\,\mathrm{PHA}_k\). When the deviation
\(\mathrm{PHA}_t - \mathrm{PHA}_{\min}\) exceeds a threshold \(\lambda_{\mathrm{PHA}}\), a significant change is declared. Formally,
\begin{equation}
    \mathrm{PHA}_t - \mathrm{PHA}_{\min} \;>\;\lambda_{\mathrm{PHA}}\;\;\;\implies\;\;\;\text{non-stationary drift detected.}
\end{equation}
In stable periods (no obvious drift), \(\rho\) remains close to \(\rho_{\min}\), exhibiting a gentle linear increase. When a significant drift is detected, $\rho$ will directly increase to $\rho_{\max}$. Concretely, at the $t$-th round:
\begin{equation}
    \rho \;=\;
    \begin{cases}
    \rho_{\max}, & \text{if } \mathrm{PHA}_t - \mathrm{PHA}_{\min} > \lambda_{\mathrm{PHA}},\\
    \rho_{\min} \;+\;\alpha\cdot (\mathrm{PHA}_t - \mathrm{PHA}_{\min}), & \text{otherwise},
    \end{cases}
\end{equation}
where \(\alpha\) is a scaling factor that modulates how quickly \(\rho\) responds to incremental changes in stable periods. Here, we carefully design the parameters such that $\alpha\cdot\lambda_{\mathrm{PHA}} \le \rho_{\max} - \rho_{\min}$, ensuring that $\rho$ will not exceed the maximum limit in the linear adjustment phase.

Before delving into the formal description, we briefly illustrate how SeRe can be seamlessly integrated into a generic CNB framework. Recall that at each round, a standard CNB algorithm typically updates the parameters of user-level (or cluster-level) models, performs clustering based on the model updates, and then uses a UCB policy to select an action for each user. 

SeRe extends this process by introducing selective reinitialization for network units that exhibit low contribution utilities. Specifically, after the CNB algorithms update each network's parameters and performs the usual clustering and arm selection, we use a change-detection mechanism to adaptively adjust the replacement rate \(\rho\). We then invoke the SeRe on updated networks, selectively reinitializing their weights when certain units meet the replacement criteria. This ensures that the network maintains sufficient plasticity in non-stationary environments, avoiding overfitting to outdated user preferences. 
Algorithm~\ref{alg:CNB_SeRe_Template} presents a high-level template that demonstrates how these modules --- CNB algorithms' general operations, adaptive change detection and SeRe --- are woven together.

\begin{algorithm}[t]
\caption{SeRe-enhanced CNB}
\label{alg:CNB_SeRe_Template}
\begin{algorithmic}[1]
\State \textbf{Input:}
\(\rho_{\min}\) (minimum replacement rate), 
\(\rho_{\max}\) (maximum replacement rate), 
\(\delta\) (Page-Hinkley-Absolute offset), 
\(\lambda_{\mathrm{PHA}}\) (Page-Hinkley-Absolute threshold),
\(\alpha\) (scaling factor)

\State \textbf{Initialize:} \(\rho \leftarrow \rho_{\min}\)
\For{all neural networks \(f\)}
    \State \quad \(u_{l, i} = 0\), \(\textit{age}_{l, i} = 0\) for all units in \(f\)
    \State \quad \(c_l = 0\) for all layers \(l\)
    \State \quad \(\mathrm{PHA}_0 = 0\), \(\mathrm{PHA}_{\min} = 0\)
\EndFor

\For{each round \(t = 1, 2, \dots\)}    
    \State \Comment{/* CNB algorithms' general operations */}
    \State \parbox[t]{\dimexpr\linewidth-\algorithmicindent}{Receive a target user $u_t \in \mathcal{N}$, observe $K$ arms $\mathcal{A}_t = \{\bm{a}_t^1, \bm{a}_t^2, \dots, \bm{a}_t^K\}$}
    \State \parbox[t]{\dimexpr\linewidth-\algorithmicindent}{Cluster users according to Definition \ref{def:cluster}}
    \State Update all relevant cluster learners
    \State \parbox[t]{\dimexpr\linewidth-\algorithmicindent}{$U^i_t$ = $\hat{r}^i_t$ + Confidence Term (calculated according to the design of specific CNB algorithms)}
    \State \parbox[t]{\dimexpr\linewidth-\algorithmicindent}{$I_t$ = $\arg\max_{i\in[K]}U^i_t$}
    \State Play \(\bm{a}^{I_t}_t\) and observe actual reward \(r^{I_t}_t\)
    \State Update all relevant user learners
    \State \Comment{/* Adaptive change detection */}
    \State Update \(\mathrm{PHA}_t \leftarrow \mathrm{PHA}_{t-1} + \bigl[\, |r^{I_t}_t - \hat{r}^{I_t}_t| - \delta \,\bigr]\)
    \State Update \(\mathrm{PHA}_{\min} \leftarrow \min(\mathrm{PHA}_{\min}, \mathrm{PHA}_t)\)
    \If{\(\mathrm{PHA}_t - \mathrm{PHA}_{\min} > \lambda_{\mathrm{PHA}}\)}
        \State \(\rho \leftarrow \rho_{\max}\)
    \Else
        \State \(\rho \leftarrow \rho_{\min} + \alpha \cdot (\mathrm{PHA}_t - \mathrm{PHA}_{\min})\)
    \EndIf
    \For{all updated \(f\) in $t$-th round}
        Algorithm \ref{alg:SeRe}
    \EndFor
\EndFor
\end{algorithmic}
\end{algorithm}

\section{Theoretical Analysis}
\label{sec:theo}

In this section, we analyze the advantages of SeRe theoretically. We analyze SeRe in the piecewise-stationary setting~\cite{appel1983adaptive, yu2009piecewise}, a widely used assumption in bandit theory to approximate non-stationary environments, serving as a theoretical bridge between dynamic regret (which considers arbitrary changes) and static regret. Under this assumption, the environment of each piece is relatively stable, but the reward function \(g_{u,t}(\cdot)\) changes abruptly at each \(\tau_s\). Let
\(
  1 = \tau_0 < \tau_1 < \dots < \tau_S < \tau_{S+1} \;=\; T+1,
\)
and assume that within each piece \(s \in \{0,1,\dots,S\}\), the reward function \(g_{u,t}(\cdot)\) for user \(u_t\) at round \(t \in [\tau_s,\tau_{s+1})\) is relatively stationary, i.e., it can be approximately regarded as
\(
  g_{u_t,t}(\bm{a}) = g_{u_t,t'}(\bm{a}), 
  \forall\, t,t' \in [\tau_s,\tau_{s+1}), \bm{a}\in \mathcal{A}.
\)

Splitting the umulative dynamic regret \(\mathbf{R}_T\) accordingly, we write
\(
  \mathbf{R}_T=
  \sum_{s=0}^S \mathbf{R}_s,
\)
where
\(
\mathbf{R}_s =
\sum_{t=\tau_s}^{\tau_{s+1}-1} 
\Bigl[g_{u_t,t}(\bm{a}_t^*) - g_{u_t,t}(\bm{a}^{I_t}_t)\Bigr]
\)
is the regret incurred within the piece $s$.
Let \(\rho_s \in (0,1)\) denote the overall fraction of units reinitialized in piece \(s\). Specifically, we regard all reinitialization performed within a piece \(s\) as simultaneously reinitializing the \(\rho_s\) fraction of units that contribute the smallest utility at its beginning. Assume that within piece \(s\), there are \(P\) learners (user-level or cluster-level) that are selectively initialized; we denote the resulting post-reset parameters by \(\widetilde{\theta}^s_p\) for \(p \in [P] := \{1, 2, \dots, P\}\). We want to show that SeRe guarantees that the updated parameters remain close (in the \(\ell_2\)-norm) to a random initialization, thereby avoiding the issues of stale or dead units. Based on this, we can propose the following theorem:

\begin{table*}[htbp]
  \centering
  \renewcommand{\arraystretch}{0.95} 
  \caption{Average regret (smaller is better) per round comparison among CLB algorithms, CNB algorithms (some of them are the neural version of CLB algorithms), and SeRe-enhanced CNB algorithms over six experiments on online recommendation datasets: the ``-N'' suffix indicates the neural version of the method, and `` + SeRe'' means this method is combined with our SeRe. In each group, if the difference between the neural version and SeRe-enhanced version is statistically significant (p $<$ 0.05), the results of SeRe-enhanced version are shown in bold with a superscript $^{*}$.}
  \label{tab:methods}
  \resizebox{\textwidth}{!}{
  \begin{tabular}{lcccccc}
    \toprule
    \textbf{Algorithm} & \textbf{KuaiRec} & \textbf{Yelp} & \textbf{MovieLens} & \textbf{Facebook} & \textbf{Amazon-Video Games} & \textbf{Amazon-Digital Music} \\
    \midrule
    \textbf{CLUB}              & 0.8104 $\pm$ 0.0025 & 0.8221 $\pm$ 0.0015 & 0.5052 $\pm$ 0.0050 & 0.5969 $\pm$ 0.0028 & 0.7376 $\pm$ 0.0013 & 0.6778 $\pm$ 0.0039 \\
    \textbf{CLUB-N}            & 0.7999 $\pm$ 0.0023 & 0.8113 $\pm$ 0.0017 & 0.4935 $\pm$ 0.0042 & 0.5846 $\pm$ 0.0015 & 0.7188 $\pm$ 0.0020 & 0.6659 $\pm$ 0.0032 \\
    \textbf{CLUB-N + SeRe}     & \textbf{0.7807 $\pm$ 0.0014}$^{*}$ & \textbf{0.7927 $\pm$ 0.0015}$^{*}$ & \textbf{0.4370 $\pm$ 0.0031}$^{*}$ & \textbf{0.5229 $\pm$ 0.0011}$^{*}$ & \textbf{0.6712 $\pm$ 0.0018}$^{*}$ & \textbf{0.6231 $\pm$ 0.0045}$^{*}$ \\
    \midrule
    \textbf{SCLUB}             & 0.7822 $\pm$ 0.0032 & 0.7931 $\pm$ 0.0014 & 0.5588 $\pm$ 0.0066 & 0.5677 $\pm$ 0.0033 & 0.6951 $\pm$ 0.0030 & 0.6103 $\pm$ 0.0069 \\
    \textbf{SCLUB-N}           & 0.7754 $\pm$ 0.0030 & 0.7921 $\pm$ 0.0008 & 0.5502 $\pm$ 0.0082 & 0.5593 $\pm$ 0.0024 & 0.6829 $\pm$ 0.0031 & 0.5916 $\pm$ 0.0041 \\
    \textbf{SCLUB-N + SeRe}    & \textbf{0.7603 $\pm$ 0.0019}$^{*}$ & \textbf{0.7760 $\pm$ 0.0012}$^{*}$ & \textbf{0.4872 $\pm$ 0.0026}$^{*}$ & \textbf{0.5179 $\pm$ 0.0028}$^{*}$ & \textbf{0.6481 $\pm$ 0.0024}$^{*}$ & \textbf{0.5505 $\pm$ 0.0010}$^{*}$ \\
    \midrule
    \textbf{LOCB}              & 0.7802 $\pm$ 0.0030 & 0.7787 $\pm$ 0.0015 & 0.5245 $\pm$ 0.0060 & 0.5446 $\pm$ 0.0088 & 0.6603 $\pm$ 0.0022 & 0.7224 $\pm$ 0.0077 \\
    \textbf{LOCB-N}            & 0.7754 $\pm$ 0.0028 & 0.7708 $\pm$ 0.0007 & 0.5188 $\pm$ 0.0037 & 0.5388 $\pm$ 0.0067 & 0.6475 $\pm$ 0.0017 & 0.7067 $\pm$ 0.0043 \\
    \textbf{LOCB-N + SeRe}     & \textbf{0.7576 $\pm$ 0.0020}$^{*}$ & \textbf{0.7549 $\pm$ 0.0006}$^{*}$ & \textbf{0.4770 $\pm$ 0.0011}$^{*}$ & \textbf{0.4872 $\pm$ 0.0156}$^{*}$ & \textbf{0.5831 $\pm$ 0.0019}$^{*}$ & \textbf{0.6763 $\pm$ 0.0035}$^{*}$ \\
    \midrule
    \textbf{M-CNB}             & 0.7146 $\pm$ 0.0044 & 0.7547 $\pm$ 0.0006 & 0.4450 $\pm$ 0.0050 & 0.3080 $\pm$ 0.0103 & 0.6357 $\pm$ 0.0041 & 0.5367 $\pm$ 0.0044 \\
    \textbf{M-CNB + SeRe}      & \textbf{0.6929 $\pm$ 0.0021}$^{*}$ & \textbf{0.7349 $\pm$ 0.0005}$^{*}$ & \textbf{0.4056 $\pm$ 0.0028}$^{*}$ & \textbf{0.2686 $\pm$ 0.0081}$^{*}$ & \textbf{0.5607 $\pm$ 0.0029}$^{*}$ & \textbf{0.4811 $\pm$ 0.0008}$^{*}$ \\
    \bottomrule
  \end{tabular}
  }
\end{table*}

\begin{theorem}[Regret Upper Bound]
\label{thm:SeRe_sublinear_main}
For SeRe-enhanced CNB algorithms, in the piecewise-stationary setting of \(S\) pieces, the cumulative dynamic
regret over \(T\) rounds satisfies
\[
\mathbf{R}_T = \sum_{s=0}^S \mathbf{R}_s = \widetilde{\mathcal{O}}\Bigl(\sqrt{T S}\Bigr).
\]
In particular, if the number of pieces satisfies \(S = o(T)\) (i.e., \(S\) grows slower than \(T\)), then the overall regret is sublinear in \(T\).
\end{theorem}

\begin{remark}
\label{rmk:theorem}
Theorem \ref{thm:SeRe_sublinear_main} ensures that the cumulative dynamic
regret remains sublinear in \(T\) as long as the number of change points \(S\) grows slower than \(T\), i.e., \(S = o(T)\). In many real-world systems, the number of major environmental shifts is relatively small compared to the total number of rounds \(T\). For example, user preferences in a large-scale recommender system do not necessarily change drastically in every round; instead, significant shifts (e.g., seasonality or major trend changes) occur only occasionally. Therefore, our theoretical proof provides valuable insights or real-world tasks.
\end{remark}

\textbf{Discussion.} 
We define \(\Delta_s = \tau_{s+1} - \tau_s\) as the length of piece \(s\). Prior analyses of neural bandits and CNB algorithms have shown that if model parameters remain within \(\mathcal{O}(M^{1/4})\) (in \(\ell_2\)-norm) of a random initialization—where \(M\) denotes the total number of units—then a UCB policy can achieve a regret bound of \(\widetilde{\mathcal{O}}(\sqrt{\Delta_s})\) in each stationary piece~\cite{ban2024meta}. 
The remaining task is to demonstrate that SeRe effectively reinitializes the network parameters of each user or cluster to a ``fresh'' region near the random initialization with high probability at every change point \(\tau_s\). This reinitialization prevents the network from inheriting suboptimal local minima from the previous piece, allowing it to quickly resume sublinear regret accumulation in the new piece. 
A detailed proof of Theorem~\ref{thm:SeRe_sublinear_main} is provided in Appendix \ref{appendix:SeRe_proofs}.

\section{Experiments}
\label{sec:experiments}
In this section, we evaluate the effectiveness of SeRe on six different online recommendation datasets.
Our code and settings are publicly available at \href{https://github.com/zhiyuansu0326/SeRe}{\textit{https://github.com/zhiyuansu0326/SeRe}} .

\textbf{Datasets.} 
We conduct experiments on six diverse online recommendation datasets: \textbf{KuaiRec}~\cite{gao2022kuairec}, which contains 7,176 users, 10,728 items, and \(1.2\times10^7\) interactions; \textbf{Yelp}\footnote{\url{https://www.yelp.com/dataset}}, with about \(7\times10^6\) reviews and \(1.5\times10^5\) business attributes; \textbf{MovieLens}~\cite{harper2015movielens}, featuring 25 million ratings and one million tag applications for $6.2\times10^4$ movies by $1.6 \times 10^5$ users; \textbf{Facebook}~\cite{leskovec2012learning}, a snapshot of 88,234 social links; and two Amazon subsets~\cite{hou2024bridging}—\textbf{Amazon - Video Games}, with 4.6 million reviews from 2.8 million users on $1.4 \times 10^5$ items, and \textbf{Amazon - Digital Music}, comprising \(1.3\times10^5\) reviews from $1\times10^5$ users on $7 \times 10^4$ items.
We first extract user ratings from raw reviews and construct a rating matrix by selecting the top 10,000 users and items (those with the most ratings). We then apply singular value decomposition (SVD) to obtain feature vectors for each user and item, normalized for consistency. A rating above 4 yields a reward of 1, otherwise 0. We further cluster users into 50 groups via K-means. In each iteration, a user is randomly sampled from one of these clusters, and a set of 10 items is formed: one item with a positive rating (reward=1) and nine with negative ratings (reward=0). These items serve as arms in a contextual bandit, where the user–item feature vectors provide the context for selecting the arm that maximizes reward.

\begin{figure*}[htbp]
    \centering
    \includegraphics[width=1\linewidth]{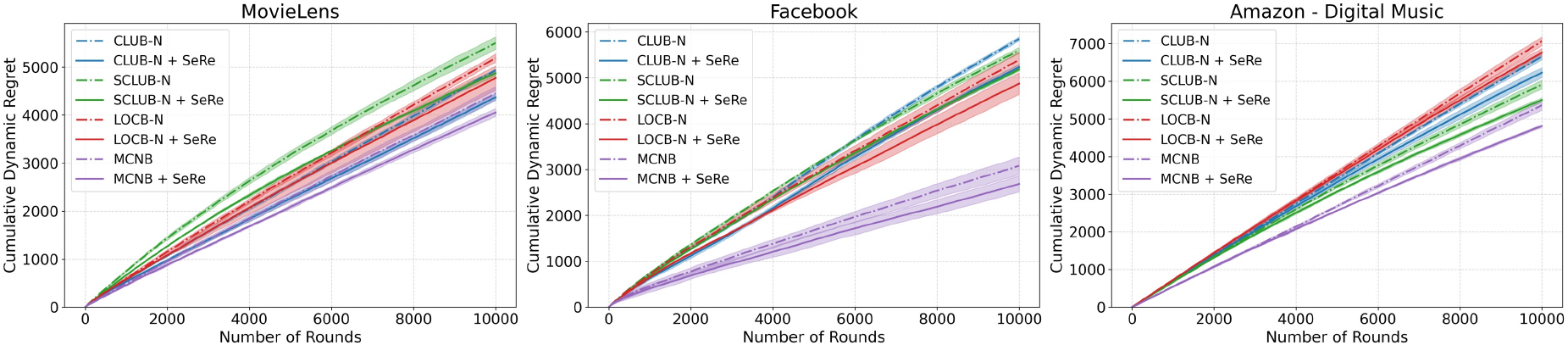}
    \caption{Regret comparison between CNB algorithms and SeRe-enhanced CNB algorithms on three online recommendation datasets: the ``-N'' suffix indicates the neural version of the method, and `` + SeRe'' means this method is combined with SeRe. Five experiments were performed for each setting: the middle line represents the average curve and the shaded area represents the 95\% confidence interval.}
    \label{fig:some_regrets}
\end{figure*}

\textbf{Baselines.} 
We claim that SeRe is an algorithm module that enhances CNB algorithms. Accordingly, we selected four SOTA CNB algorithms as baselines. In Section \ref{sec:loss_of_plasticity}, we detailed our neural extension of three existing CLB algorithms into CNB algorithms, and we additionally include the M-CNB algorithm as a baseline. Our baselines are: (1) \textbf{CLUB-N}, extended from CLUB~\cite{gentile2014online}, which adaptively clusters users via evolving similarity graphs and connected components in an online contextual bandit setting, thereby balancing exploration and exploitation with regret bounds based on cluster geometry and size; (2) \textbf{SCLUB-N}, extended from SCLUB~\cite{li2019improved}, which employs a split-and-merge strategy with set-based clustering to handle non-uniform user frequencies and remove dependencies on minimal frequency probabilities; (3) \textbf{LOCB-N}, extended from LOCB~\cite{ban2021local}, a local clustering algorithm that incrementally refines clusters starting from seeds, supports overlapping clusters, and provides theoretical guarantees on clustering efficiency and regret; and (4) \textbf{M-CNB}~\cite{ban2024meta}, which combines meta-learning with neural networks to dynamically adapt user clusters based on arbitrary reward functions, using a UCB-based exploration policy with improved theoretical and empirical performance. 

\textbf{Results.} Table~\ref{tab:methods} shows the average per-round regret for CLB algorithms, CNB algorithms (some of them are the neural version of CLB algorithms), and SeRe-enhanced CNB algorithms across six online recommendation datasets. We observed that when comparing CNB algorithms with their SeRe-enhanced counterparts, all the models incorporating SeRe show statistically significant improvements (p $<$ 0.05) over those without SeRe across all datasets. Through all the experimental data, SeRe can reduce the average regret by 12.82\% in 10,000 rounds at most (M-CNB and M-CNB + SeRe on Facebook dataset). In addition, we selected the cumulative dynamic regret curves on three datasets for display in Figure \ref{fig:some_regrets}, and the full regret figure on six datasets are attached at Figure \ref{fig:regrets} in Appendix~\ref{appendix:all_regrets}. The results show that the regrets of the CNB algorithms combined with SeRe are generally lower than its corresponding baseline algorithms, and show obvious sublinear accumulation. The most obvious improvement of SeRe is shown on Amazon-Digital Music, the curve of SCLUB-N almost degenerates into linearity, but the curve of SCLUB-N + SeRe clearly maintains a good sublinear trend.
This indicates that SeRe effectively enhances CNB algorithms by mitigating the loss of plasticity and improving adaptation to dynamic, non-stationary recommendation.

\textbf{Runtime Analysis.}  
In streaming recommendation scenarios, runtime is a critical performance metric. In Table \ref{tab:runtime_increase}, we evaluate the average additional runtime incurred by integrating SeRe into our CNB algorithms on the two largest datasets, KuaiRec and Yelp. In our experiments, SeRe adds only a few milliseconds per round to the online model training and recommendation process. Considering the real-time constraints of streaming recommendations and the outstanding performance improvement (regret reduction) of up to 12.82\%, this additional delay is negligible and acceptable. These results confirm that SeRe can significantly reduce accumulated regret and enhance model plasticity and adaptability with very little additional running cost, which is a great advantage.

\begin{table}[t]
    \centering
    \renewcommand{\arraystretch}{0.90}
    \caption{Additional runtime (in millisecond/round) for CNB algorithms integrated with SeRe (mean ± standard deviation).}
    \label{tab:runtime_increase}
    \resizebox{\columnwidth}{!}{  
    \begin{tabular}{lcccc}
        \toprule
        \textbf{Dataset} & \textbf{CLUB-N} & \textbf{SCLUB-N} & \textbf{LOCB-N} & \textbf{M-CNB} \\
        \midrule
        \textbf{KuaiRec} & $2.52 \pm 0.15$ & $2.38 \pm 0.10$ & $2.94 \pm 0.12$ & $3.07 \pm 0.08$ \\
        \textbf{Yelp}    & $2.82 \pm 0.24$ & $2.66 \pm 0.18$ & $3.18 \pm 0.16$ & $3.36 \pm 0.10$ \\
        \bottomrule
    \end{tabular}
    }
\end{table}

\begin{figure}[htbp]
    \centering
    \includegraphics[width=1\linewidth]{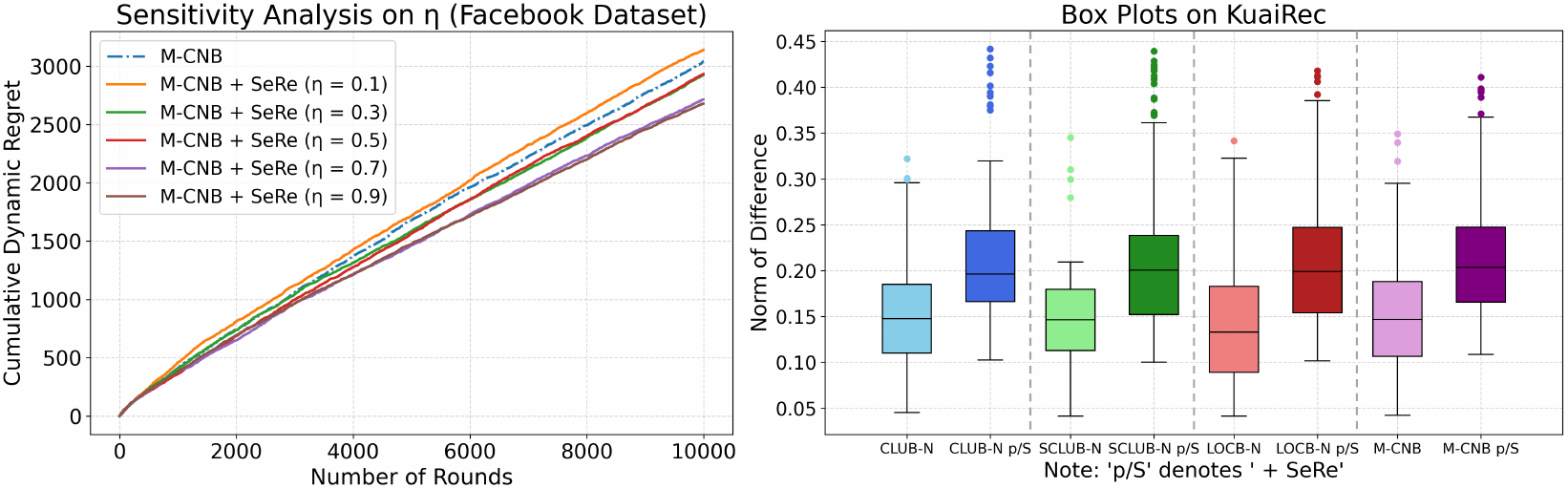}
    \caption{Sensitivity and Plasticity Analysis. (1)~The left panel: comparison of regret curves of MCNB and MCNB + SeRe under different \(\eta\). (2)~The right panel: this box plot illustrates the $\ell_2$-norm of the difference in the last layer's parameters, computed from samples taken every 25 rounds over 10,000 rounds on the KuaiRec dataset.}
    \label{fig:sensitivity}
\end{figure}

\textbf{Sensitivity and Plasticity Analysis.}
In Figure \ref{fig:sensitivity}, we perform sensitivity and plasticity analysis. 
In the left panel, we perform a sensitivity analysis on \(\eta\) (decay rate) using the Facebook dataset with the M-CNB and M-CNB + SeRe models.
Our experiments over \(\eta\) values (0.1, 0.3, 0.5, 0.7, 0.9) reveal that the lowest cumulative regret is achieved at \(\eta = 0.9\) followed by \(\eta = 0.7\), whereas \(\eta = 0.1\) leads to inferior performance, even worse than the baseline.
This can be explained from Update Equation~\eqref{eq:contribution_utility_time_agnostic}: a larger \(\eta\) gives greater weight to past contribution utility, yielding smoother updates, whereas a smaller \(\eta\) causes more abrupt fluctuations, which can destabilize learning and hinder SeRe's effectiveness in adapting to both non-stationary and stationary environments. In addition, the sensitivity analysis of \(m\) (maturity threshold) can be found in Appendix \ref{appendix:sens_m}. Note that the hyperparameters mentioned in Section \ref{sec:sere_app} are not part of the Algorithm \ref{alg:SeRe}, do not directly determine the unit contribution utility update and maturity judgment, and have limited impact on the timing and effect of reinitialization triggering, so we can use grid search to match the most appropriate value for a specific task. The range of the grid search is listed in the Appendix \ref{appendix:other_parameters}.
The right panel illustrates the norm of the difference of the last layer's parameters every 25 rounds over 10,000 rounds on KuaiRec dataset, which exhibits strong temporal variation. After combining with SeRe (right side of each group), the median and maximum values of the norm are significantly improved compared to the baselines (left side of each group). This shows that SeRe allows the parameters of the models to change to a greater extent in a non-stationary environment, which makes the models more plastic.
\vspace{-1em}

\vspace{-1em}

\textbf{Empirical Analysis of Reinitialization Frequency.} Although the frequency of reinitialization in SeRe varies depending on the degree of user preference shift across different datasets, our adaptive change detection mechanism ensures that it remains selective and infrequent. We conducted an empirical analysis using the KuaiRec dataset to quantify this behavior. The interval between reinitialization rounds ranged from 26 to 47 steps, and only 2.6\% to 7.2\% of all rounds involved any reinitialization across different CNB algorithms. These results indicate that SeRe reinitializes only a small number of units at controlled intervals, avoiding excessive disruption to the learned model. Moreover, this reinitialization behavior remains relatively consistent across SeRe-enhanced CNB algorithms, confirming the stability and reliability of our mechanism in dynamic environments.

\section{Conclusion}
\label{sec:conclusion}
In the context of Clustering of Neural Bandits (CNB), loss of plasticity refers to the reduced adaptability of neural network parameters over time, which poses a significant challenge in dynamic, non-stationary environments. 
To address this issue, we propose Selective Reinitialization (SeRe), a model-agnostic and continual bandit learning framework that selectively resets underutilized units to maintain flexibility while preserving accumulated knowledge. By dynamically adjusting reinitialization frequency to environmental changes, SeRe enables CNB algorithms to achieve sublinear cumulative dynamic regret. Empirical results confirm its effectiveness in improving adaptability and reducing regret, making SeRe a practical solution for sustaining model plasticity in dynamic settings.

\section*{Acknowledgments}
This work was partially supported by the National Natural Science Foundation of China (No. 62376275, 62472426). Work partially done at Beijing Key Laboratory of Research on Large Models and Intelligent Governance, and Engineering Research Center of Next-Generation Intelligent Search and Recommendation, MOE. Supported by fund for building world-class universities (disciplines) of Renmin University of China.

\bibliographystyle{ACM-Reference-Format}
\bibliography{KDD25}

@String{Computer = "{IEEE} Computer" }

@String{Springer = "Springer-Verlag" }

@inproceedings{li2010contextual,
  title={A contextual-bandit approach to personalized news article recommendation},
  author={Li, Lihong and Chu, Wei and Langford, John and Schapire, Robert E},
  booktitle={Proceedings of the 19th international conference on World wide web},
  pages={661--670},
  year={2010}
}

@inproceedings{ban2021local,
  title={Local clustering in contextual multi-armed bandits},
  author={Ban, Yikun and He, Jingrui},
  booktitle={Proceedings of the Web Conference 2021},
  pages={2335--2346},
  year={2021}
}

@inproceedings{Zhang2022Counteracting,
  title={Counteracting user attention bias in music streaming recommendation via reward modification},
  author={Zhang, Xiao and Dai, Sunhao and Xu, Jun and Dong, Zhenhua and Dai, Quanyu and Wen, Ji-Rong},
  booktitle={Proceedings of the 28th ACM SIGKDD Conference on Knowledge Discovery and Data Mining},
  pages={2504--2514},
  year={2022}
}

@inproceedings{mehrotra2020bandit,
  title={Bandit based optimization of multiple objectives on a music streaming platform},
  author={Mehrotra, Rishabh and Xue, Niannan and Lalmas, Mounia},
  booktitle={Proceedings of the 26th ACM SIGKDD international conference on knowledge discovery \& data mining},
  pages={3224--3233},
  year={2020}
}

@inproceedings{gentile2014online,
  title={Online clustering of bandits},
  author={Gentile, Claudio and Li, Shuai and Zappella, Giovanni},
  booktitle={International conference on machine learning},
  pages={757--765},
  year={2014},
  organization={PMLR}
}

@inproceedings{zhang2025adao2b,
  title={AdaO2B: Adaptive Online to Batch Conversion for Out-of-Distribution Generalization},
  author={Zhang, Xiao and Dai, Sunhao and Xu, Jun and Liu, Yong and Dong, Zhenhua},
  booktitle={Proceedings of the AAAI Conference on Artificial Intelligence},
  volume={39},
  number={21},
  pages={22596--22604},
  year={2025}
}

@inproceedings{yang2016tracking,
  title={Tracking slowly moving clairvoyant: Optimal dynamic regret of online learning with true and noisy gradient},
  author={Yang, Tianbao and Zhang, Lijun and Jin, Rong and Yi, Jinfeng},
  booktitle={International Conference on Machine Learning},
  pages={449--457},
  year={2016},
  organization={PMLR}
}

@inproceedings{gentile2017context,
  title={On context-dependent clustering of bandits},
  author={Gentile, Claudio and Li, Shuai and Kar, Purushottam and Karatzoglou, Alexandros and Zappella, Giovanni and Etrue, Evans},
  booktitle={International Conference on machine learning},
  pages={1253--1262},
  year={2017},
  organization={PMLR}
}

@article{wang2024online,
  title={Online clustering of bandits with misspecified user models},
  author={Wang, Zhiyong and Xie, Jize and Liu, Xutong and Li, Shuai and Lui, John},
  journal={Advances in Neural Information Processing Systems},
  volume={36},
  year={2024}
}

@inproceedings{zhang2024reinforcing,
  title={Reinforcing Long-Term Performance in Recommender Systems with User-Oriented Exploration Policy},
  author={Zhang, Changshuo and Chen, Sirui and Zhang, Xiao and Dai, Sunhao and Yu, Weijie and Xu, Jun},
  booktitle={Proceedings of the 47th International ACM SIGIR Conference on Research and Development in Information Retrieval},
  pages={1850--1860},
  year={2024}
}

@inproceedings{li2016collaborative,
  title={Collaborative filtering bandits},
  author={Li, Shuai and Karatzoglou, Alexandros and Gentile, Claudio},
  booktitle={Proceedings of the 39th International ACM SIGIR conference on Research and Development in Information Retrieval},
  pages={539--548},
  year={2016}
}

@inproceedings{ban2024meta,
  title={Meta clustering of neural bandits},
  author={Ban, Yikun and Qi, Yunzhe and Wei, Tianxin and Liu, Lihui and He, Jingrui},
  booktitle={Proceedings of the 30th ACM SIGKDD Conference on Knowledge Discovery and Data Mining},
  pages={95--106},
  year={2024}
}

@inproceedings{zhou2020neural,
  title={Neural contextual bandits with ucb-based exploration},
  author={Zhou, Dongruo and Li, Lihong and Gu, Quanquan},
  booktitle={International Conference on Machine Learning},
  pages={11492--11502},
  year={2020},
  organization={PMLR}
}

@article{zhang2023reward,
  title={Reward imputation with sketching for contextual batched bandits},
  author={Zhang, Xiao and Shao, Ninglu and Si, Zihua and Xu, Jun and Wang, Wenhan and Su, Hanjing and Wen, Ji-Rong},
  journal={Advances in Neural Information Processing Systems 36},
  pages={64577--64588},
  year={2023}
}

@inproceedings{zhang2021counterfactual,
  title={Counterfactual reward modification for streaming recommendation with delayed feedback},
  author={Zhang, Xiao and Jia, Haonan and Su, Hanjing and Wang, Wenhan and Xu, Jun and Wen, Ji-Rong},
  booktitle={Proceedings of the 44th international ACM SIGIR conference on research and development in information retrieval},
  pages={41--50},
  year={2021}
}

@article{dohare2024loss,
  title={Loss of plasticity in deep continual learning},
  author={Dohare, Shibhansh and Hernandez-Garcia, J Fernando and Lan, Qingfeng and Rahman, Parash and Mahmood, A Rupam and Sutton, Richard S},
  journal={Nature},
  volume={632},
  number={8026},
  pages={768--774},
  year={2024},
  publisher={Nature Publishing Group UK London}
}

@article{li2019improved,
  title={Improved algorithm on online clustering of bandits},
  author={Li, Shuai and Chen, Wei and Leung, Kwong-Sak},
  journal={arXiv preprint arXiv:1902.09162},
  year={2019}
}

@article{valko2013finite,
  title={Finite-time analysis of kernelised contextual bandits},
  author={Valko, Michal and Korda, Nathaniel and Munos, R{\'e}mi and Flaounas, Ilias and Cristianini, Nelo},
  journal={arXiv preprint arXiv:1309.6869},
  year={2013}
}

@article{xu2020neural,
  title={Neural contextual bandits with deep representation and shallow exploration},
  author={Xu, Pan and Wen, Zheng and Zhao, Handong and Gu, Quanquan},
  journal={arXiv preprint arXiv:2012.01780},
  year={2020}
}

@article{cybenko1989approximation,
  title={Approximation by superpositions of a sigmoidal function},
  author={Cybenko, George},
  journal={Mathematics of control, signals and systems},
  volume={2},
  number={4},
  pages={303--314},
  year={1989},
  publisher={Springer}
}

@inproceedings{zhao2020simple,
  title={A simple approach for non-stationary linear bandits},
  author={Zhao, Peng and Zhang, Lijun and Jiang, Yuan and Zhou, Zhi-Hua},
  booktitle={International Conference on Artificial Intelligence and Statistics},
  pages={746--755},
  year={2020},
  organization={PMLR}
}

@article{kumar2023maintaining,
  title={Maintaining Plasticity in Continual Learning via Regenerative Regularization},
  author={Kumar, Saurabh and Marklund, Henrik and Van Roy, Benjamin},
  year={2023}
}

@article{van2022three,
  title={Three types of incremental learning},
  author={Van de Ven, Gido M and Tuytelaars, Tinne and Tolias, Andreas S},
  journal={Nature Machine Intelligence},
  volume={4},
  number={12},
  pages={1185--1197},
  year={2022},
  publisher={Nature Publishing Group UK London}
}

@article{page1954continuous,
  title={Continuous inspection schemes},
  author={Page, Ewan S},
  journal={Biometrika},
  volume={41},
  number={1/2},
  pages={100--115},
  year={1954},
  publisher={JSTOR}
}

@article{hinkley1971inference,
  title={Inference about the change-point from cumulative sum tests},
  author={Hinkley, David V},
  journal={Biometrika},
  volume={58},
  number={3},
  pages={509--523},
  year={1971},
  publisher={Oxford University Press}
}

@inproceedings{allen2019convergence,
  title={A convergence theory for deep learning via over-parameterization},
  author={Allen-Zhu, Zeyuan and Li, Yuanzhi and Song, Zhao},
  booktitle={International conference on machine learning},
  pages={242--252},
  year={2019},
  organization={PMLR}
}

@inproceedings{cao2020generalization,
  title={Generalization error bounds of gradient descent for learning over-parameterized deep relu networks},
  author={Cao, Yuan and Gu, Quanquan},
  booktitle={Proceedings of the AAAI Conference on Artificial Intelligence},
  volume={34},
  number={04},
  pages={3349--3356},
  year={2020}
}

@inproceedings{gao2022kuairec,
  author = {Gao, Chongming and Li, Shijun and Lei, Wenqiang and Chen, Jiawei and Li, Biao and Jiang, Peng and He, Xiangnan and Mao, Jiaxin and Chua, Tat-Seng},
  title = {KuaiRec: A Fully-Observed Dataset and Insights for Evaluating Recommender Systems},
  booktitle = {Proceedings of the 31st ACM International Conference on Information \& Knowledge Management},
  series = {CIKM '22},
  location = {Atlanta, GA, USA},
  url = {https://doi.org/10.1145/3511808.3557220},
  doi = {10.1145/3511808.3557220},
  numpages = {11},
  year = {2022},
  pages = {540–550}
}

@article{leskovec2012learning,
  title={Learning to discover social circles in ego networks},
  author={Leskovec, Jure and Mcauley, Julian},
  journal={Advances in neural information processing systems},
  volume={25},
  year={2012}
}

@article{harper2015movielens,
  title={The movielens datasets: History and context},
  author={Harper, F Maxwell and Konstan, Joseph A},
  journal={Acm transactions on interactive intelligent systems (tiis)},
  volume={5},
  number={4},
  pages={1--19},
  year={2015},
  publisher={Acm New York, NY, USA}
}

@article{hou2024bridging,
  title={Bridging Language and Items for Retrieval and Recommendation},
  author={Hou, Yupeng and Li, Jiacheng and He, Zhankui and Yan, An and Chen, Xiusi and McAuley, Julian},
  journal={arXiv preprint arXiv:2403.03952},
  year={2024}
}

@article{filippi2010parametric,
  title={Parametric bandits: The generalized linear case},
  author={Filippi, Sarah and Cappe, Olivier and Garivier, Aur{\'e}lien and Szepesv{\'a}ri, Csaba},
  journal={Advances in neural information processing systems},
  volume={23},
  year={2010}
}

@article{soare2014best,
  title={Best-arm identification in linear bandits},
  author={Soare, Marta and Lazaric, Alessandro and Munos, R{\'e}mi},
  journal={Advances in Neural Information Processing Systems},
  volume={27},
  year={2014}
}

@article{zhang2020neural,
  title={Neural thompson sampling},
  author={Zhang, Weitong and Zhou, Dongruo and Li, Lihong and Gu, Quanquan},
  journal={arXiv preprint arXiv:2010.00827},
  year={2020}
}

@article{jia2022learning,
  title={Learning neural contextual bandits through perturbed rewards},
  author={Jia, Yiling and Zhang, Weitong and Zhou, Dongruo and Gu, Quanquan and Wang, Hongning},
  journal={arXiv preprint arXiv:2201.09910},
  year={2022}
}

@article{ban2022improved,
  title={Improved algorithms for neural active learning},
  author={Ban, Yikun and Zhang, Yuheng and Tong, Hanghang and Banerjee, Arindam and He, Jingrui},
  journal={Advances in Neural Information Processing Systems},
  volume={35},
  pages={27497--27509},
  year={2022}
}

@article{ban2024neural,
  title={Neural active learning beyond bandits},
  author={Ban, Yikun and Agarwal, Ishika and Wu, Ziwei and Zhu, Yada and Weldemariam, Kommy and Tong, Hanghang and He, Jingrui},
  journal={arXiv preprint arXiv:2404.12522},
  year={2024}
}

@article{qi2024meta,
  title={Meta-learning with neural bandit scheduler},
  author={Qi, Yunzhe and Ban, Yikun and Wei, Tianxin and Zou, Jiaru and Yao, Huaxiu and He, Jingrui},
  journal={Advances in Neural Information Processing Systems},
  volume={36},
  year={2024}
}

@inproceedings{santana2020contextual,
  title={Contextual meta-bandit for recommender systems selection},
  author={Santana, Marlesson RO and Melo, Luckeciano C and Camargo, Fernando HF and Brand{\~a}o, Bruno and Soares, Anderson and Oliveira, Renan M and Caetano, Sandor},
  booktitle={Proceedings of the 14th ACM Conference on Recommender Systems},
  pages={444--449},
  year={2020}
}

@inproceedings{kassraie2022neural,
  title={Neural contextual bandits without regret},
  author={Kassraie, Parnian and Krause, Andreas},
  booktitle={International Conference on Artificial Intelligence and Statistics},
  pages={240--278},
  year={2022},
  organization={PMLR}
}

@article{hong2020latent,
  title={Latent bandits revisited},
  author={Hong, Joey and Kveton, Branislav and Zaheer, Manzil and Chow, Yinlam and Ahmed, Amr and Boutilier, Craig},
  journal={Advances in Neural Information Processing Systems},
  volume={33},
  pages={13423--13433},
  year={2020}
}

@inproceedings{maillard2014latent,
  title={Latent Bandits.},
  author={Maillard, Odalric-Ambrym and Mannor, Shie},
  booktitle={International Conference on Machine Learning},
  pages={136--144},
  year={2014},
  organization={PMLR}
}

@inproceedings{shen2023hyperbandit,
  title={HyperBandit: Contextual Bandit with Hypernewtork for Time-Varying User Preferences in Streaming Recommendation},
  author={Shen, Chenglei and Zhang, Xiao and Wei, Wei and Xu, Jun},
  booktitle={Proceedings of the 32nd ACM International Conference on Information and Knowledge Management},
  pages={2239--2248},
  year={2023}
}

@inproceedings{korda2016distributed,
  title={Distributed clustering of linear bandits in peer to peer networks},
  author={Korda, Nathan and Szorenyi, Balazs and Li, Shuai},
  booktitle={International conference on machine learning},
  pages={1301--1309},
  year={2016},
  organization={PMLR}
}

@inproceedings{liu2022federated,
  title={Federated online clustering of bandits},
  author={Liu, Xutong and Zhao, Haoru and Yu, Tong and Li, Shuai and Lui, John CS},
  booktitle={Uncertainty in Artificial Intelligence},
  pages={1221--1231},
  year={2022},
  organization={PMLR}
}

@article{robins1995catastrophic,
  title={Catastrophic forgetting, rehearsal and pseudorehearsal},
  author={Robins, Anthony},
  journal={Connection Science},
  volume={7},
  number={2},
  pages={123--146},
  year={1995},
  publisher={Taylor \& Francis}
}

@article{kirkpatrick2017overcoming,
  title={Overcoming catastrophic forgetting in neural networks},
  author={Kirkpatrick, James and Pascanu, Razvan and Rabinowitz, Neil and Veness, Joel and Desjardins, Guillaume and Rusu, Andrei A and Milan, Kieran and Quan, John and Ramalho, Tiago and Grabska-Barwinska, Agnieszka and others},
  journal={Proceedings of the national academy of sciences},
  volume={114},
  number={13},
  pages={3521--3526},
  year={2017},
  publisher={National Acad Sciences}
}

@article{lopez2017gradient,
  title={Gradient episodic memory for continual learning},
  author={Lopez-Paz, David and Ranzato, Marc'Aurelio},
  journal={Advances in neural information processing systems},
  volume={30},
  year={2017}
}

@article{shin2017continual,
  title={Continual learning with deep generative replay},
  author={Shin, Hanul and Lee, Jung Kwon and Kim, Jaehong and Kim, Jiwon},
  journal={Advances in neural information processing systems},
  volume={30},
  year={2017}
}

@article{ellis2000age,
  title={Age of acquisition effects in adult lexical processing reflect loss of plasticity in maturing systems: insights from connectionist networks.},
  author={Ellis, Andrew W and Lambon Ralph, Matthew A},
  journal={Journal of Experimental Psychology: Learning, memory, and cognition},
  volume={26},
  number={5},
  pages={1103},
  year={2000},
  publisher={American Psychological Association}
}

@inproceedings{chaudhry2018riemannian,
  title={Riemannian walk for incremental learning: Understanding forgetting and intransigence},
  author={Chaudhry, Arslan and Dokania, Puneet K and Ajanthan, Thalaiyasingam and Torr, Philip HS},
  booktitle={Proceedings of the European conference on computer vision (ECCV)},
  pages={532--547},
  year={2018}
}

@article{dohare2023maintaining,
  title={Maintaining plasticity in deep continual learning},
  author={Dohare, Shibhansh and Hernandez-Garcia, J Fernando and Rahman, Parash and Mahmood, A Rupam and Sutton, Richard S},
  journal={arXiv preprint arXiv:2306.13812},
  year={2023}
}

@inproceedings{sokar2023dormant,
  title={The dormant neuron phenomenon in deep reinforcement learning},
  author={Sokar, Ghada and Agarwal, Rishabh and Castro, Pablo Samuel and Evci, Utku},
  booktitle={International Conference on Machine Learning},
  pages={32145--32168},
  year={2023},
  organization={PMLR}
}

@inproceedings{abbas2023loss,
  title={Loss of plasticity in continual deep reinforcement learning},
  author={Abbas, Zaheer and Zhao, Rosie and Modayil, Joseph and White, Adam and Machado, Marlos C},
  booktitle={Conference on Lifelong Learning Agents},
  pages={620--636},
  year={2023},
  organization={PMLR}
}

@article{appel1983adaptive,
  title={Adaptive sequential segmentation of piecewise stationary time series},
  author={Appel, Ulrich and Brandt, Achim V},
  journal={Information sciences},
  volume={29},
  number={1},
  pages={27--56},
  year={1983},
  publisher={Elsevier}
}

@inproceedings{yu2009piecewise,
  title={Piecewise-stationary bandit problems with side observations},
  author={Yu, Jia Yuan and Mannor, Shie},
  booktitle={Proceedings of the 26th annual international conference on machine learning},
  pages={1177--1184},
  year={2009}
}

@inproceedings{zhang2018dynamic,
  title={Dynamic regret of strongly adaptive methods},
  author={Zhang, Lijun and Yang, Tianbao and Zhou, Zhi-Hua and others},
  booktitle={International conference on machine learning},
  pages={5882--5891},
  year={2018},
  organization={PMLR}
}

@article{fei2020dynamic,
  title={Dynamic regret of policy optimization in non-stationary environments},
  author={Fei, Yingjie and Yang, Zhuoran and Wang, Zhaoran and Xie, Qiaomin},
  journal={Advances in Neural Information Processing Systems},
  volume={33},
  pages={6743--6754},
  year={2020}
}

@article{zhou2020nonstationary,
  title={Nonstationary reinforcement learning with linear function approximation},
  author={Zhou, Huozhi and Chen, Jinglin and Varshney, Lav R and Jagmohan, Ashish},
  journal={arXiv preprint arXiv:2010.04244},
  year={2020}
}

@article{adel2019continual,
  title={Continual learning with adaptive weights (claw)},
  author={Adel, Tameem and Zhao, Han and Turner, Richard E},
  journal={arXiv preprint arXiv:1911.09514},
  year={2019}
}

@inproceedings{kurle2019continual,
  title={Continual learning with bayesian neural networks for non-stationary data},
  author={Kurle, Richard and Cseke, Botond and Klushyn, Alexej and Van Der Smagt, Patrick and G{\"u}nnemann, Stephan},
  booktitle={International Conference on Learning Representations},
  year={2019}
}

@inproceedings{wang2022meta,
  title={Meta-learning with less forgetting on large-scale non-stationary task distributions},
  author={Wang, Zhenyi and Shen, Li and Fang, Le and Suo, Qiuling and Zhan, Donglin and Duan, Tiehang and Gao, Mingchen},
  booktitle={European Conference on Computer Vision},
  pages={221--238},
  year={2022},
  organization={Springer}
}

@inproceedings{liu2020incremental,
  title={Incremental few-shot meta-learning via indirect discriminant alignment},
  author={Liu, Qing and Majumder, Orchid and Achille, Alessandro and Ravichandran, Avinash and Bhotika, Rahul and Soatto, Stefano},
  booktitle={Computer Vision--ECCV 2020: 16th European Conference, Glasgow, UK, August 23--28, 2020, Proceedings, Part VII 16},
  pages={685--701},
  year={2020},
  organization={Springer}
}

@inproceedings{rudner2022continual,
  title={Continual learning via sequential function-space variational inference},
  author={Rudner, Tim GJ and Smith, Freddie Bickford and Feng, Qixuan and Teh, Yee Whye and Gal, Yarin},
  booktitle={International Conference on Machine Learning},
  pages={18871--18887},
  year={2022},
  organization={PMLR}
}

@inproceedings{tseran2018natural,
  title={Natural variational continual learning},
  author={Tseran, Hanna and Khan, Mohammad Emtiyaz and Harada, Tatsuya and Bui, Thang D},
  booktitle={Continual Learning Workshop@ NeurIPS},
  volume={2},
  year={2018}
}

@article{miao2022context,
  title={Context-based dynamic pricing with online clustering},
  author={Miao, Sentao and Chen, Xi and Chao, Xiuli and Liu, Jiaxi and Zhang, Yidong},
  journal={Production and Operations Management},
  volume={31},
  number={9},
  pages={3559--3575},
  year={2022},
  publisher={SAGE Publications Sage CA: Los Angeles, CA}
}

@article{wang2025dynamic,
  title={On dynamic pricing with covariates},
  author={Wang, Hanzhao and Talluri, Kalyan and Li, Xiaocheng},
  journal={Operations Research},
  year={2025},
  publisher={INFORMS}
}

@article{wang2023adcb,
  title={ADCB: Adaptive Dynamic Clustering of Bandits for Online Recommendation System},
  author={Wang, Yufeng and Zhang, Weidong and Ma, Jianhua and Jin, Qun},
  journal={Neural Processing Letters},
  volume={55},
  number={2},
  pages={1155--1172},
  year={2023},
  publisher={Springer}
}

@inproceedings{li2018online,
  title={Online clustering of contextual cascading bandits},
  author={Li, Shuai and Zhang, Shengyu},
  booktitle={Proceedings of the AAAI Conference on Artificial Intelligence},
  volume={32},
  number={1},
  year={2018}
}

@inproceedings{nguyen2014dynamic,
  title={Dynamic clustering of contextual multi-armed bandits},
  author={Nguyen, Trong T and Lauw, Hady W},
  booktitle={Proceedings of the 23rd ACM international conference on conference on information and knowledge management},
  pages={1959--1962},
  year={2014}
}

@inproceedings{cheung2019learning,
  title={Learning to optimize under non-stationarity},
  author={Cheung, Wang Chi and Simchi-Levi, David and Zhu, Ruihao},
  booktitle={The 22nd International Conference on Artificial Intelligence and Statistics},
  pages={1079--1087},
  year={2019},
  organization={PMLR}
}

@inproceedings{he2015delving,
  title={Delving deep into rectifiers: Surpassing human-level performance on imagenet classification},
  author={He, Kaiming and Zhang, Xiangyu and Ren, Shaoqing and Sun, Jian},
  booktitle={Proceedings of the IEEE international conference on computer vision},
  pages={1026--1034},
  year={2015}
}

@article{ban2021ee,
  title={Ee-net: Exploitation-exploration neural networks in contextual bandits},
  author={Ban, Yikun and Yan, Yuchen and Banerjee, Arindam and He, Jingrui},
  journal={arXiv preprint arXiv:2110.03177},
  year={2021}
}

@article{dai2022federated,
  title={Federated neural bandits},
  author={Dai, Zhongxiang and Shu, Yao and Verma, Arun and Fan, Flint Xiaofeng and Low, Bryan Kian Hsiang and Jaillet, Patrick},
  journal={arXiv preprint arXiv:2205.14309},
  year={2022}
}

\appendix
\section{Detailed Theoretical Analysis and Proofs}
\label{appendix:SeRe_proofs}

In this appendix, we provide a rigorous derivation of the regret upper bound for the SeRe-enhanced CNB algorithm. Our analysis rests on the theoretical framework of Neural Contextual Bandits in the \textit{Over-parameterized Regime} (Neural Tangent Kernel regime), specifically extending the results of \citet{zhou2020neural} and \citet{ban2024meta} to the piecewise-stationary setting.

\subsection{Preliminaries and Problem Setup}

Consider a time horizon $T$. The environment is piecewise-stationary with $S$ stationary segments. The change points are denoted by $\mathcal{T} = \{ \tau_0, \tau_1, \dots, \tau_S, \tau_{S+1} \}$, where $\tau_0 = 1$ and $\tau_{S+1} = T+1$. The $s$-th stationary segment is defined as $\mathcal{I}_s = [\tau_s, \tau_{s+1} - 1]$ with length $H_s = \tau_{s+1} - \tau_s$.

We define the filtration $\mathfrak{F}_{t-1} = \sigma(\{u_\tau, \bm{a}_\tau, r_\tau\}_{\tau=1}^{t-1} \cup \{u_t, \mathcal{A}_t\})$ representing the history up to round $t$. The dynamic regret is defined as:
\begin{equation}
    \mathbf{R}_T = \sum_{t=1}^T \left( g_{u_t, t}(\bm{a}_t^*) - g_{u_t, t}(\bm{a}_t^{I_t}) \right).
\end{equation}

\subsection{Regularity Assumptions}

To ensure the tractability of the neural bandit optimization, we adopt standard assumptions from the NTK literature.

\textbf{(a) Boundedness.}
For all $t \in [T]$, the true reward function satisfies $g_{u,t}(\bm{a}) \in [0, 1]$. The feature vectors satisfy $\|\bm{a}\|_2 \leq 1$. The noise $\xi_t$ is 1-sub-Gaussian conditioned on $\mathfrak{F}_{t-1}$.

\textbf{(b) Over-parameterized Neural Network.}
Let $f(x; \theta)$ be the neural network with width $m$ and depth $L$. We assume $m$ is sufficiently large, satisfying $m \geq \text{poly}(T, L, 1/\delta)$, such that with probability at least $1-\delta$, for all $t$, the trained parameters $\theta_t$ satisfy $\|\theta_t - \theta_0\|_2 \leq R$, where $R$ is a constant radius. Within this ball, the network can be approximated by its first-order Taylor expansion (linearization) with approximation error bounded by $\mathcal{O}(m^{-1/2})$.

\textbf{(c) Bounded Detection Delay.}
Let $\hat{\tau}_s$ denote the time step when the SeRe mechanism (via Page-Hinkley-Absolute test) triggers a reinitialization corresponding to the true change point $\tau_s$. We assume there exists a bounded delay constant $\Delta > 0$ such that $0 \le \hat{\tau}_s - \tau_s \le \Delta$. During the interval $[\tau_s, \hat{\tau}_s)$, the algorithm may suffer linear regret.

\textbf{(d) Effective Reinitialization.}
Upon reinitialization at $\hat{\tau}_s$, the reinitialized parameters $\tilde{\theta}_{\hat{\tau}_s}$ are drawn from the initialization distribution $\mathcal{D}$. We assume this reinitialization restores the random feature properties required for Assumption (b) to hold for the new reward function $g_{u, t}$ in the segment $\mathcal{I}_s$. Specifically, the neural exploration bonus becomes valid relative to the new ground truth after reinitialization.

\subsection{Key Lemma: Regret in Stationary Neural Bandits}

We first restate the regret bound for standard CNB/NeuralUCB in a \textit{single} stationary environment.

\begin{lemma}[Stationary Regret Bound~\cite{ban2024meta}]
\label{lemma:stationary}
Consider a stationary period of length $H$. Under Assumptions (a) and (b), with probability at least $1-\delta'$, the cumulative regret of the CNB algorithm is bounded by:
\begin{equation}
    R_{\text{static}}(H, \delta') \leq C \cdot \sqrt{\tilde{d} H \log(H / \delta')},
\end{equation}
where $\tilde{d}$ is the effective dimension of the Neural Tangent Kernel matrix, and $C$ is a constant depending on network depth $L$ and radius $R$.
\end{lemma}

\subsection{Proof of Theorem \ref{thm:SeRe_sublinear_main}}

\begin{proof}
The total time horizon $T$ is partitioned into $S$ true stationary segments plus the adaptation delays. For each change point $\tau_s$ (where $s=1,\dots,S$), the SeRe mechanism detects the change at $\hat{\tau}_s$. This splits the learning process into two types of intervals:
\begin{enumerate}
    \item \textbf{Delay Intervals} $\mathcal{J}_s^{\text{delay}} = [\tau_s, \hat{\tau}_s)$: The period where the environment has changed, but the model has not yet reinitialized. Length $\leq \Delta$.
    \item \textbf{Effective Learning Intervals} $\mathcal{J}_s^{\text{learn}} = [\hat{\tau}_s, \tau_{s+1})$: The period where the model has reinitialized and learns the new distribution. Let $T_s = |\mathcal{J}_s^{\text{learn}}|$ be the effective learning length for segment $s$.
\end{enumerate}

The total regret $\mathbf{R}_T$ can be decomposed as:
\begin{equation}
    \mathbf{R}_T = \sum_{s=0}^S \left( \sum_{t \in \mathcal{J}_s^{\text{delay}}} r_t^{\text{regret}} + \sum_{t \in \mathcal{J}_s^{\text{learn}}} r_t^{\text{regret}} \right).
\end{equation}

\textbf{Step 1: Bounding Delay Regret.}
Under Assumption (a), the maximum regret per step is 1. Under Assumption (c), the length of each delay interval is at most $\Delta$. Thus:
\begin{equation}
    \label{eq:delay_bound}
    \mathbf{R}_{\text{delay}} = \sum_{s=1}^S \sum_{t \in \mathcal{J}_s^{\text{delay}}} r_t^{\text{regret}} \leq \sum_{s=1}^S \Delta \cdot 1 = S \Delta.
\end{equation}

\textbf{Step 2: Bounding Learning Regret (Union Bound).}
For the learning intervals, we apply Lemma \ref{lemma:stationary}. To ensure the bound holds simultaneously for all $S+1$ segments with a global high probability of at least $1-\delta$, we must set the failure probability for each individual segment to $\delta' = \delta / (S+1)$.

Applying Lemma \ref{lemma:stationary} to the $s$-th learning interval of length $T_s$:
\begin{equation}
    \text{Regret}(\mathcal{J}_s^{\text{learn}}) \leq C \sqrt{\tilde{d} T_s \log\left(\frac{T_s (S+1)}{\delta}\right)}.
\end{equation}
Let $\beta_T = C \sqrt{\tilde{d} \log\left(\frac{T (S+1)}{\delta}\right)}$. Since $T_s \leq T$, we can simplify the logarithmic factor for the summation:
\begin{equation}
    \label{eq:learn_bound_raw}
    \mathbf{R}_{\text{learn}} = \sum_{s=0}^S \text{Regret}(\mathcal{J}_s^{\text{learn}}) \leq \sum_{s=0}^S \beta_T \sqrt{T_s}.
\end{equation}

\textbf{Step 3: Aggregation via Cauchy-Schwarz.}
We maximize the sum $\sum_{s=0}^S \sqrt{T_s}$ subject to the constraint $\sum_{s=0}^S T_s \leq T$.
Consider vectors $\mathbf{u} = [1, \dots, 1]^\top \in \mathbb{R}^{S+1}$ and $\mathbf{v} = [\sqrt{T_0}, \dots, \sqrt{T_S}]^\top \in \mathbb{R}^{S+1}$. By Cauchy-Schwarz inequality ($\mathbf{u}^\top \mathbf{v} \leq \|\mathbf{u}\|_2 \|\mathbf{v}\|_2$):
\begin{equation}
    \sum_{s=0}^S \sqrt{T_s} \leq \sqrt{\sum_{s=0}^S 1^2} \cdot \sqrt{\sum_{s=0}^S (\sqrt{T_s})^2} = \sqrt{S+1} \cdot \sqrt{\sum_{s=0}^S T_s} \leq \sqrt{(S+1)T}.
\end{equation}

Substituting this back into Eq. (\ref{eq:learn_bound_raw}):
\begin{equation}
    \label{eq:learn_bound_final}
    \mathbf{R}_{\text{learn}} \leq \beta_T \sqrt{(S+1)T} = C \sqrt{\tilde{d} (S+1) T \log\left(\frac{T(S+1)}{\delta}\right)}.
\end{equation}

\textbf{Step 4: Final Bound.}
Combining Eq. (\ref{eq:delay_bound}) and Eq. (\ref{eq:learn_bound_final}), with probability at least $1-\delta$:
\begin{equation}
    \mathbf{R}_T \leq S \Delta + C \sqrt{\tilde{d} (S+1) T \log\left(\frac{T(S+1)}{\delta}\right)}.
\end{equation}

Using asymptotic notation $\widetilde{\mathcal{O}}$ to hide logarithmic factors and constants (assuming $\Delta$ is a small constant relative to $T$), and noting that $S+1 \approx S$:
\begin{equation}
    \mathbf{R}_T = \widetilde{\mathcal{O}}\left( S + \sqrt{S T} \right) = \widetilde{\mathcal{O}}\left( \sqrt{S T} \right).
\end{equation}
The last equality holds because for the regret to be sublinear, we implicitly assume $S \ll T$, making $\sqrt{ST}$ the dominant term.

This completes the proof.
\end{proof}

\subsection{Reinitialization Argument}

We now present a lemma ensuring that reinitializing a fraction \(\rho_s\) of the lowest-utility units in each layer can bring the post-reset parameter vector \(\widetilde{\theta}^{u,s}\) for user \(u\) in piece \(s\) into a ``good region'' around its fresh initialization \(\theta_0^u\); that is, 
\[
\|\widetilde{\theta}^{u,s} - \theta_0^u\|_2 \le \mathcal{O}(M^{1/4})
\]
with high probability. This property is crucial for maintaining network plasticity.

\begin{lemma}[Selective Reinitialization Preserves Freshness]
\label{lemma:selective_reinit_freshness}
Let \(\rho_s \in (0,1)\) be the fraction of units replaced at piece \(s\). Assume that the weight blocks corresponding to the replaced units have bounded expected norm and that each new unit's weights are i.i.d.\ samples from the initial distribution \(\mathcal{D}\). Then, with high probability,
\[
\|\widetilde{\theta}^{u,s} - \theta_0^u\|_2 
\;\le\; 
\omega,
\]
for some \(\omega = \mathcal{O}(M^{1/4})\), uniformly over all users \(u \in \{1,\dots,n\}\).
\end{lemma}

\begin{proof}
We aim to show that after selectively reinitializing a fraction \(\rho_s\) of units in piece \(s\), the updated (post-reset) parameter vector \(\widetilde{\theta}^{u,s}\) for each user \(u\) remains within \(\omega = \mathcal{O}(M^{1/4})\) (in \(\ell_2\)-norm) of its fresh initialization \(\theta_0^u\) with high probability. Let \(m\) be the total number of units (or parameter blocks) in the network for user \(u\), and denote the pre-reset parameter vector in piece \(s\) by
\[
\theta^{u,s} = \bigl[\theta^{u,s}_1, \theta^{u,s}_2, \dots, \theta^{u,s}_m\bigr].
\]
Suppose \(\theta_0^u\) is drawn from \(\mathcal{D}\) with
\[
\mathbb{E}\Bigl[\|\theta_0^u\|_2^2\Bigr] \le C,
\]
for some constant \(C>0\) independent of \(m\). Let \(\lfloor \rho_s m \rfloor\) units with the lowest contribution utility be replaced, with each new unit's weights independently sampled from \(\mathcal{D}\).

Define the post-reset parameter for each unit as
\[
\widetilde{\theta}^{u,s}_i = 
\begin{cases}
\theta^{u,s}_i, & \text{if unit \(i\) is retained}, \\
\text{a fresh sample from \(\mathcal{D}\)}, & \text{if unit \(i\) is reinitialized}.
\end{cases}
\]
Then the overall post-reset parameter vector is
\[
\widetilde{\theta}^{u,s} = \bigl[\widetilde{\theta}^{u,s}_1, \widetilde{\theta}^{u,s}_2, \dots, \widetilde{\theta}^{u,s}_m\bigr],
\]
and we have
\[
\|\widetilde{\theta}^{u,s} - \theta_0^u\|_2^2
\;=\;
\sum_{i=1}^m \Bigl\|\widetilde{\theta}^{u,s}_i - \theta^u_{0,i}\Bigr\|_2^2.
\]
Partition the index set \(\{1,\dots,m\}\) into 
\(\mathcal{S}_{\text{replaced}}\) (with \(|\mathcal{S}_{\text{replaced}}| = \lfloor \rho_s m \rfloor\)) and 
\(\mathcal{S}_{\text{retained}}\) (the remaining indices). Thus,
\[
\|\widetilde{\theta}^{u,s} - \theta_0^u\|_2^2
\;=\;
\sum_{i \in \mathcal{S}_{\text{replaced}}} 
\Bigl\|\widetilde{\theta}^{u,s}_i - \theta^u_{0,i}\Bigr\|_2^2
\;+\;
\sum_{i \in \mathcal{S}_{\text{retained}}}
\Bigl\|\theta^{u,s}_i - \theta^u_{0,i}\Bigr\|_2^2.
\]

For each \(i \in \mathcal{S}_{\text{replaced}}\), since \(\widetilde{\theta}^{u,s}_i\) is an independent sample from \(\mathcal{D}\), we have
\[
\mathbb{E}\Bigl[\Bigl\|\widetilde{\theta}^{u,s}_i - \theta^u_{0,i}\Bigr\|_2^2\Bigr] 
=\mathbb{E}\Bigl[\|\widetilde{\theta}^{u,s}_i\|_2^2 + \|\theta^u_{0,i}\|_2^2\Bigr] 
\;\approx\; 2\,\mathbb{E}\Bigl[\|\theta^u_{0,i}\|_2^2\Bigr] 
\;\le\; 2C.
\]
Summing over all replaced units yields
\[
\mathbb{E}\!\Biggl[\sum_{i \in \mathcal{S}_{\text{replaced}}}
\Bigl\|\widetilde{\theta}^{u,s}_i - \theta^u_{0,i}\Bigr\|_2^2\Biggr]
\;\le\; 2\,\lfloor \rho_s m \rfloor \cdot C
\;\approx\; 2\,\rho_s\,m\,C.
\]

By a standard concentration argument (e.g., via Hoeffding's or Bernstein's inequality) and choosing an appropriate deviation term, we obtain with high probability
\[
\sum_{i \in \mathcal{S}_{\text{replaced}}}
\Bigl\|\widetilde{\theta}^{u,s}_i - \theta^u_{0,i}\Bigr\|_2^2
\;\le\; \mathcal{O}(\rho_s\,M).
\]

For each \(i \in \mathcal{S}_{\text{retained}}\), since \(\widetilde{\theta}^{u,s}_i = \theta^{u,s}_i\), we have
\[
\Bigl\|\widetilde{\theta}^{u,s}_i - \theta^u_{0,i}\Bigr\|_2^2 
=\Bigl\|\theta^{u,s}_i - \theta^u_{0,i}\Bigr\|_2^2.
\]
By assumption (or by bounding the network’s drift over time), we assume that
\[
\sum_{i \in \mathcal{S}_{\text{retained}}}
\Bigl\|\theta^{u,s}_i - \theta^u_{0,i}\Bigr\|_2^2
\;\le\; (1 - \rho_s)\,M \cdot \mathcal{O}\Bigl(\frac{1}{M}\Bigr)
\;=\; \mathcal{O}(1-\rho_s).
\]

Combining these bounds, we obtain
\[
\|\widetilde{\theta}^{u,s} - \theta_0^u\|_2^2 
\;\le\; \mathcal{O}(\rho_s\,M) + \mathcal{O}(1-\rho_s)
\;=\; \mathcal{O}(\rho_s\,M).
\]
Choosing \(\rho_s \sim \frac{1}{\sqrt{M}}\) makes the right-hand side \(\mathcal{O}(\sqrt{M})\), so that
\[
\|\widetilde{\theta}^{u,s} - \theta_0^u\|_2 
\;\le\; \mathcal{O}(M^{1/4}).
\]
Applying a union bound over all \(n\) users yields that, with high probability, 
\[
\|\widetilde{\theta}^{u,s} - \theta_0^u\|_2 \le \omega = \mathcal{O}(M^{1/4})
\]
for every user \(u \in \{1,\dots,n\}\). This proves that selective reinitialization successfully preserves freshness for all users simultaneously.
\end{proof}

Typically, one analyzes each replaced parameter block (in each layer) to ensure that the resulting \(\|\widetilde{\theta}^{u,s} - \theta_0^u\|_2\) remains within the desired \(\omega = \mathcal{O}(M^{1/4})\). This guarantees that the users' networks return to a ``fresh'' region near their random initializations.

\subsection{Within-piece Analysis}

Consider a stationary piece \(s\) during which the reward function \(g_{u,t}(\cdot)\) remains fixed for all rounds \(t \in [\tau_s, \tau_{s+1})\). Let \(\Delta_s = \tau_{s+1} - \tau_s\) denote the length of piece \(s\). In this setting, we assume that for each user \(u\), the post-reset parameter vector \(\widetilde{\theta}^{u,s}\) (obtained after selective reinitialization at the beginning of piece \(s\)) satisfies
\[
\|\widetilde{\theta}^{u,s} - \theta_0^u\|_2 \le \omega,
\]
with \(\omega = \mathcal{O}(M^{1/4})\). Here, \(m\) denotes the total number of units (or parameter blocks) in the network, which reflects the network's width.

We now state a key lemma that provides a UCB-based regret bound for each piece under this ``freshness'' condition.

\begin{lemma}[UCB Bound for Freshly Initialized Parameters]
\label{lemma:ucb_piece_bound}
Suppose that for each user \(u\), the post-reset parameter vector \(\widetilde{\theta}^{u,s}\) satisfies
\[
\|\widetilde{\theta}^{u,s} - \theta_0^u\|_2 \le \omega,
\]
with \(\omega = \mathcal{O}(M^{1/4})\). In piece \(s\), assume that the algorithm runs a UCB-based policy that uses \(\widetilde{\theta}^{u,s}\) as the effective initial parameters for estimation. If the reward function \(g_{u,t}(\cdot)\) remains stationary on \([\tau_s, \tau_{s+1})\), then the cumulative regret in piece \(s\) satisfies
\[
\mathbf{R}_s = \sum_{t=\tau_s}^{\tau_{s+1}-1} \Bigl[ \max_{\bm{a}\in \mathcal{A}_t} g_{u,t}(\bm{a}) - g_{u,t}\bigl(\bm{a}_t^{I_t}\bigr) \Bigr]
\le \widetilde{\mathcal{O}}(\sqrt{\Delta_s}),
\]
where \(\widetilde{\mathcal{O}}\) hides polylogarithmic factors.
\end{lemma}

\begin{proof}
For each user \(u\) in piece \(s\), let 
\[
\hat{g}_{u,t}(\bm{a}) = f\bigl(\bm{a}; \widetilde{\theta}^{u,s}\bigr)
\]
denote the model's prediction at round \(t\) (which remains fixed throughout piece \(s\)). Since \(\widetilde{\theta}^{u,s}\) remains close to the fresh initialization \(\theta_0^u\) (i.e., \(\|\widetilde{\theta}^{u,s} - \theta_0^u\|_2 \le \omega\) with \(\omega = \mathcal{O}(M^{1/4})\)), we can invoke overparameterized neural-bandit (NTK) arguments (see, e.g.,~\cite{allen2019convergence, zhou2020neural, cao2020generalization}). In particular, with high probability, the prediction error satisfies
\[
\bigl| g_{u,t}(\bm{a}) - \hat{g}_{u,t}(\bm{a}) \bigr| \le \beta_t\, \|\bm{a}\|_2,
\]
with a confidence radius \(\beta_t = \mathcal{O}\bigl(\omega \sqrt{\log t}\bigr)\).

Let \(\bm{a}_t^*\) denote the optimal arm in \(\mathcal{A}_t\) (i.e., \(\bm{a}_t^* = \arg\max_{\bm{a}\in \mathcal{A}_t} g_{u,t}(\bm{a})\)) and let \(\bm{a}_t^{I_t}\) denote the arm selected by the algorithm at round \(t\). We decompose the instantaneous regret as
\[
\begin{aligned}
g_{u,t}(\bm{a}_t^*) - g_{u,t}\bigl(\bm{a}_t^{I_t}\bigr)
  = &\underbrace{\Bigl[\hat{g}_{u,t}(\bm{a}_t^*) - \hat{g}_{u,t}(\bm{a}_t^{I_t})\Bigr]}_{\text{model difference}}+\\ &\underbrace{\Bigl(\bigl[g_{u,t}(\bm{a}_t^*) - \hat{g}_{u,t}(\bm{a}_t^*)\bigr] - \bigl[g_{u,t}(\bm{a}_t^{I_t}) - \hat{g}_{u,t}(\bm{a}_t^{I_t})\bigr]\Bigr)}_{\text{approximation error}}.
\end{aligned}
\]
Since
\[
\bigl| g_{u,t}(\bm{a}) - \hat{g}_{u,t}(\bm{a}) \bigr| \le \beta_t \|\bm{a}\|_2,
\]
we obtain
\[
g_{u,t}(\bm{a}_t^*) - g_{u,t}\bigl(\bm{a}_t^{I_t}\bigr)
\le \Bigl[\hat{g}_{u,t}(\bm{a}_t^*) - \hat{g}_{u,t}(\bm{a}_t^{I_t})\Bigr] + 2\,\beta_t\,\max_{\bm{a}\in \mathcal{A}_t}\|\bm{a}\|_2.
\]
By the design of the UCB-based policy, the selected arm \(\bm{a}_t^{I_t}\) satisfies
\[
\hat{g}_{u,t}(\bm{a}_t^{I_t}) \ge \hat{g}_{u,t}(\bm{a}_t^*) - \gamma_t,
\]
with an exploration bonus \(\gamma_t = \mathcal{O}\bigl(\sqrt{\log(t)/t}\bigr)\). Consequently,
\[
g_{u,t}(\bm{a}_t^*) - g_{u,t}\bigl(\bm{a}_t^{I_t}\bigr)
\le 2\,\beta_t\,\max_{\bm{a}\in \mathcal{A}_t}\|\bm{a}\|_2 + \gamma_t.
\]
Summing over rounds \(t = \tau_s, \dots, \tau_{s+1}-1\) in piece \(s\) gives
\[
\mathbf{R}_s = \sum_{t=\tau_s}^{\tau_{s+1}-1} \Bigl[g_{u,t}(\bm{a}_t^*) - g_{u,t}\bigl(\bm{a}_t^{I_t}\bigr)\Bigr]
\le \sum_{t=\tau_s}^{\tau_{s+1}-1} \Bigl[2\,\beta_t\,\max_{\bm{a}\in \mathcal{A}_t}\|\bm{a}\|_2 + \gamma_t\Bigr].
\]
Standard neural-bandit analysis allows us to bound \(\max_{\bm{a}\in \mathcal{A}_t}\|\bm{a}\|_2\) by a constant, so that
\[
\sum_{t=\tau_s}^{\tau_{s+1}-1} \beta_t = \mathcal{O}\bigl(\omega\,\sqrt{\Delta_s \log(\Delta_s)}\bigr)
\quad\text{and} \quad
\sum_{t=\tau_s}^{\tau_{s+1}-1} \gamma_t = \mathcal{O}\bigl(\sqrt{\Delta_s}\bigr).
\]
Therefore, the cumulative regret in piece \(s\) satisfies
\[
\mathbf{R}_s \le \widetilde{\mathcal{O}}\bigl(\sqrt{\Delta_s}\bigr),
\]
where \(\widetilde{\mathcal{O}}\) hides polylogarithmic factors in \(T\) or \(\Delta_s\).
\end{proof}

\subsection{Summation over Pieces}

\begin{proof}[Proof of Theorem \ref{thm:SeRe_sublinear_main}]
We now combine the per-piece regret bounds across all pieces \(s = 0, 1, \dots, S\). Recall that the cumulative dynamic regret is defined as
\(
\mathbf{R}_T = \sum_{s=0}^S \mathbf{R}_s.
\)
Since for each piece \(s\) we have
\(
\mathbf{R}_s =\widetilde{\mathcal{O}}\bigl(\sqrt{\Delta_s}\bigr),
\)
it follows that
\[
\mathbf{R}_T = \widetilde{\mathcal{O}}\Bigl(\sum_{s=0}^S \sqrt{\Delta_s}\Bigr).
\]
Because \(\sum_{s=0}^S \Delta_s = T\), by the Cauchy-Schwarz inequality we have
\[
\sum_{s=0}^S \sqrt{\Delta_s} \le \sqrt{(S+1)\,T},
\]
so that
\[
\mathbf{R}_T = \widetilde{\mathcal{O}}\bigl(\sqrt{TS}\bigr).
\]
Thus, if the number of pieces satisfies \(S = o(T)\) (e.g. \(S = \mathcal{O}(T^\alpha)\) for some \(\alpha < 1\)), the cumulative dynamic
regret \(\mathbf{R}_T\) grows sublinearly in \(T\).
\end{proof}

Once selective reinitialization ensures that each piece's user (or cluster) parameters are ``fresh,'' classical CNB theory implies
\[
\mathbf{R}_s = \widetilde{\mathcal{O}}\bigl(\sqrt{\Delta_s}\bigr).
\]

As mentioned in the Remark \ref{rmk:theorem}, in realistic scenarios it is natural and reasonable to assume that the number of substantial change points \(S\) grows slower than \(T\). This completes the proof of Theorem~\ref{thm:SeRe_sublinear_main}.
For more detailed proofs of Lemma~\ref{lemma:stationary}, Lemma~\ref{lemma:selective_reinit_freshness} and Lemma~\ref{lemma:ucb_piece_bound}
, please refer to~\cite{allen2019convergence,ban2024meta,zhou2020neural,cao2020generalization}.

\section{More Figures and Notes of Experiments}

\subsection{Full Regret Figure on All Datasets}
\begin{figure}[htbp]
    \centering
    \includegraphics[width=1.0\linewidth]{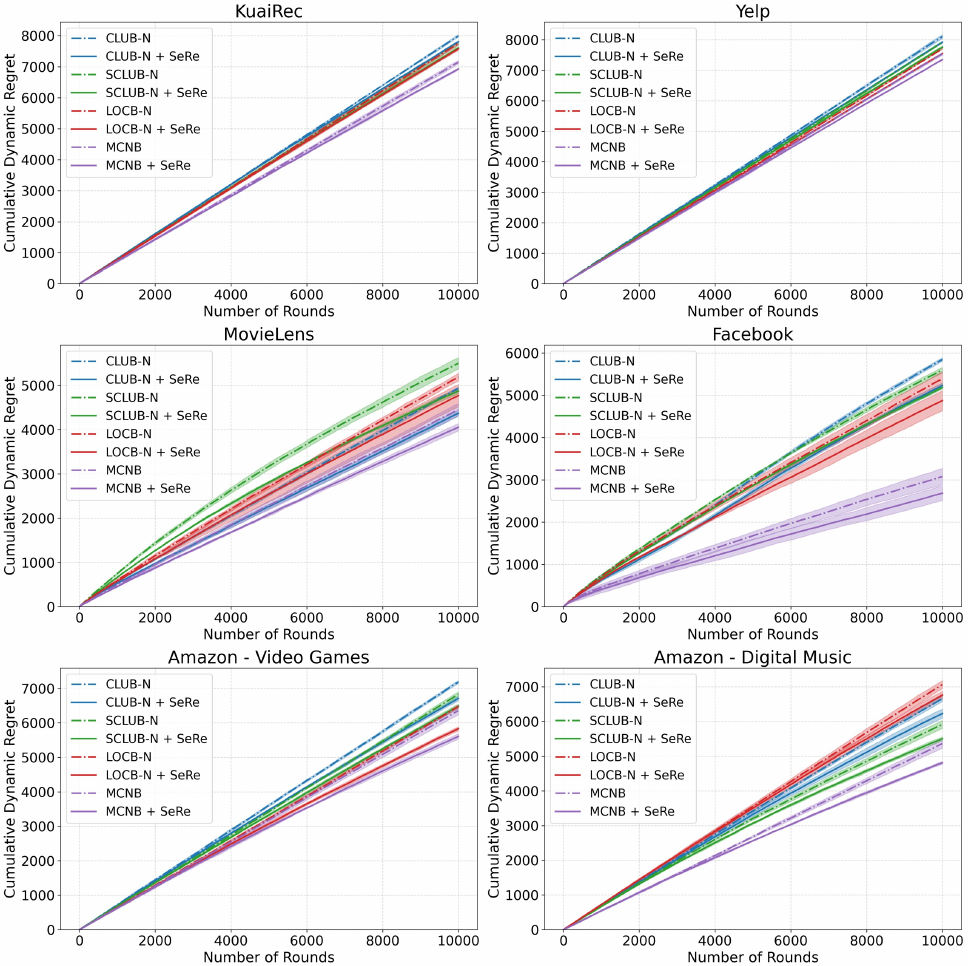}
    \caption{Regret comparison between CNB algorithms and SeRe-enhanced CNB algorithms on six online recommendation datasets: the ``-N'' suffix indicates the neural version of the method, and `` + SeRe'' means this method is combined with SeRe. Five experiments were performed for each setting: the middle line represents the average curve and the shaded area represents the 95\% confidence interval.}
    \label{fig:regrets}
\end{figure}

In this section, we will show Figure \ref{fig:regrets}, which is the full regret figure on six datasets mentioned in the ``Results'' paragraph in Section \ref{sec:experiments}.

\subsection{Sensitivity Analysis on \(m\)}
\label{appendix:sens_m}
\begin{figure}[htbp]
    \centering
    \includegraphics[width=0.8\linewidth]{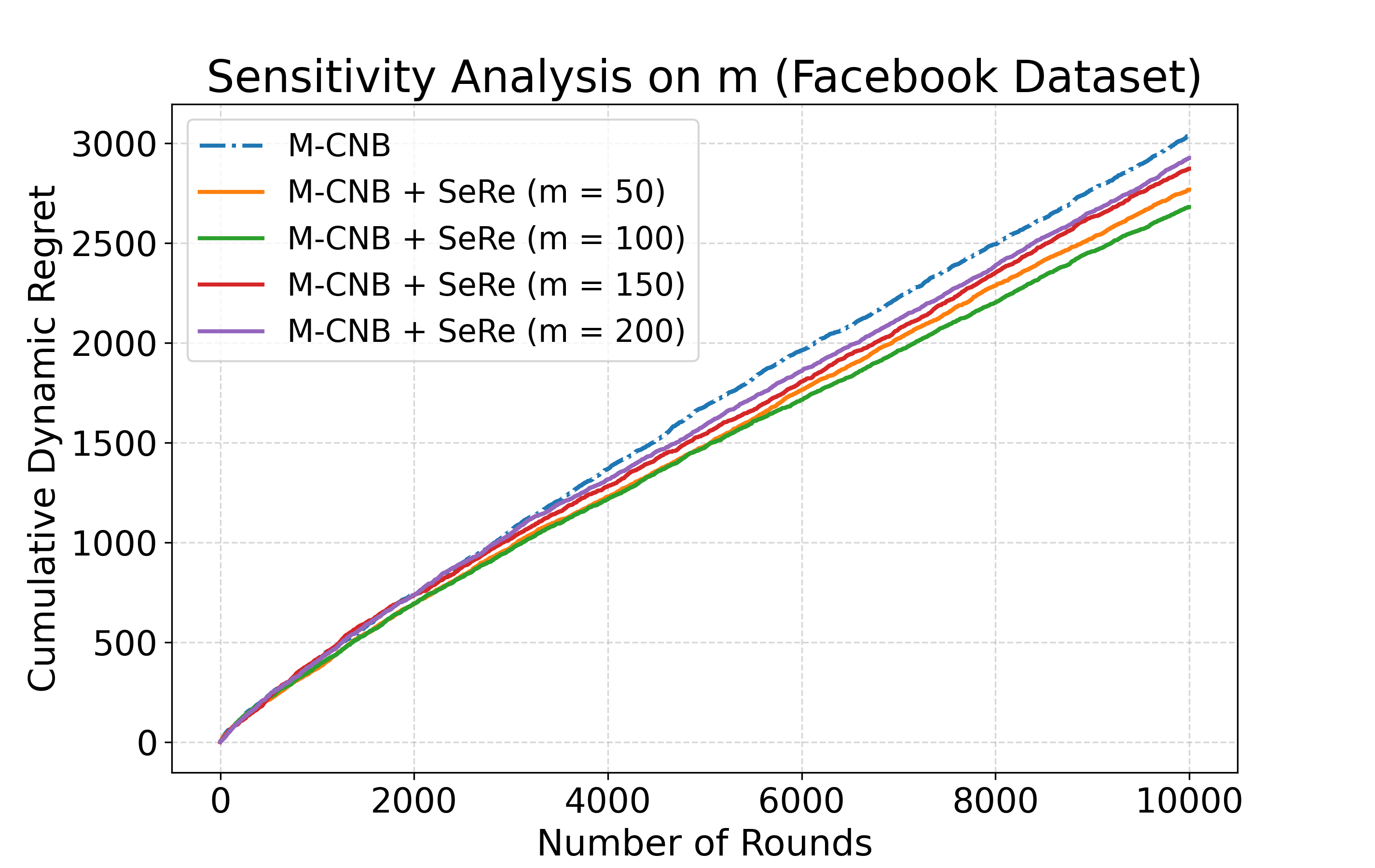}
    \caption{Comparison of regret curves of MCNB and MCNB + SeRe under different \(m\).}
    \label{fig:sens_m}
\end{figure}

In Figure \ref{fig:sens_m}, we find that regardless of the value of $m$ (50, 100, 150, 200), M-CNB + SeRe always outperforms the baseline model, and the cumulative regret is relatively less affected by changes in $m$. Additionally, we observe that an appropriate $m$ value can make SeRe’s performance improvement more pronounced, as shown in the figure where $m = 100$ achieves the lowest cumulative regret.  

\subsection{Grid Search for Other Hyperparameters}
\label{appendix:other_parameters}
In our experiments, besides the sensitivity analysis conducted on \(\eta\) and \(m\), we performed grid search for the hyperparameters mentioned in Section \ref{sec:sere_app}. Specifically, the lower replacement rate \(\rho_{\min}\) was tuned over the range \([0.005, 0.02]\) and the upper replacement rate \(\rho_{\max}\) over \([0.05, 0.2]\). We also varied the PH offset \(\delta\) in \([0.05, 0.2]\), the PH threshold \(\lambda_{\mathrm{PHA}}\) in \([0.3, 0.7]\), and the scaling factor \(\alpha\) in \([0.005, 0.02]\). These are the empirical ranges found in some of the existing research~\cite{page1954continuous,hinkley1971inference,dohare2023maintaining,dohare2024loss}. We can use grid search to match the most appropriate hyperparameter values for each specific task.

\label{appendix:all_regrets}

\vspace{-2.0em}

\end{document}